%% file: AdvCL.tex
\documentclass[10pt]{article} 
\usepackage[accepted]{tmlr}

\input{math_commands.tex}

\usepackage{hyperref}
\hypersetup{
    colorlinks=true,
    linkcolor=red,
    filecolor=magenta,      
    urlcolor=blue,
    citecolor=purple,
    pdftitle={Overleaf Example},
    pdfpagemode=FullScreen,
    }

\usepackage{epsfig}
\usepackage{graphicx}
\usepackage{amsmath}
\usepackage{amssymb}
\usepackage{multirow}
\usepackage{footmisc}
\usepackage{colortbl,booktabs}
\usepackage{threeparttable}
\usepackage{comment}
\usepackage{multicol}
\usepackage{url}
\usepackage{wrapfig}
\usepackage{mathtools}
\usepackage{pifont}
\usepackage{lipsum}
\usepackage{color,xcolor}
\definecolor{Tianlong_color}{rgb}{0.858, 0.188, 0.478}
\usepackage{adjustbox}
\usepackage{booktabs}       
\usepackage{amsfonts}       
\usepackage{nicefrac}       
\usepackage{microtype}      
\usepackage{makecell}
\usepackage{diagbox} 
\usepackage{subfig}
\usepackage[ruled, linesnumbered]{algorithm2e}

\usepackage{tikz}
\usepackage{enumitem}
\usepackage{environ}
\NewEnviron{elaboration}{
\par
\begin{tikzpicture}
\node[rectangle,minimum width=0.98\textwidth] (m) {\begin{minipage}{0.98\textwidth}\BODY\end{minipage}};
\draw[thick] (m.south west) rectangle (m.north east);
\end{tikzpicture}
}

\DeclarePairedDelimiterX{\inp}[2]{\langle}{\rangle}{#1, #2}

\DeclareMathAlphabet\mathbfcal{OMS}{cmsy}{b}{n}
\newcommand{\Def}[0]{\mathrel{\mathop:}=}

\newcommand{\RV}[1]{\textcolor{blue}{#1}}

\title{Queried Unlabeled Data Improves and Robustifies Class-Incremental Learning}

\author{\name Tianlong Chen \email tianlong.chen@utexas.edu \\
      \addr University of Texas at Austin
      \AND
      \name Sijia Liu \email liusiji5@msu.edu \\
      \addr Michigan State University \\
      MIT-IBM Watson AI Lab, IBM Research
      \AND
      \name Shiyu Chang \email chang87@ucsb.edu\\
      \addr University of California, Santa Barbara 
      \AND
      \name Lisa Amini \email lisa.amini@us.ibm.com \\
      \addr MIT-IBM Watson AI Lab, IBM Research
      \AND
      \name Zhangyang Wang \email atlaswang@utexas.edu \\
      \addr University of Texas at Austin}

\def\month{06}  
\def\year{2022} 
\def\openreview{\url{https://openreview.net/forum?id=oLvlPJheCD}} 

\begin{document}

\maketitle

\begin{abstract}
Class-incremental learning (CIL) suffers from the notorious dilemma between learning newly added classes and preserving previously learned class knowledge. That catastrophic forgetting issue could be mitigated by storing historical data for replay, which yet would cause memory overheads as well as imbalanced prediction updates. To address this dilemma, we propose to leverage ``free'' external unlabeled data querying in continual learning. We first present a CIL with Queried Unlabeled Data (\textbf{CIL-QUD}) scheme, where we only store a handful of past training samples as anchors and use them to query relevant unlabeled examples each time. Along with new and past stored data, the queried unlabeled are effectively utilized, through learning-without-forgetting (LwF) regularizers and class-balance training. Besides preserving model generalization over past and current tasks, we next study the problem of adversarial robustness for CIL-QUD.
Inspired by the recent success of learning robust models with unlabeled data, we explore a new robustness-aware CIL setting, where the learned adversarial robustness has to resist forgetting and be transferred as new tasks come in continually. While existing options easily fail, we show queried unlabeled data can continue to benefit, and seamlessly extend CIL-QUD into its robustified versions, \textbf{RCIL-QUD}. Extensive experiments demonstrate that CIL-QUD achieves substantial accuracy gains on CIFAR-10 and CIFAR-100, compared to previous state-of-the-art CIL approaches. Moreover, RCIL-QUD establishes the first strong milestone for robustness-aware CIL. Codes are available in \url{https://github.com/VITA-Group/CIL-QUD}. 
\end{abstract}

\section{Introduction}

Most deep neural networks (DNNs) are trained when the complete dataset and all class information are available at once and fixed. However, real-world applications, such as robotics and mobile health, often demand learning classifiers continually~\citep{parisi2019continual}, when the data and classes are presented and fitted sequentially. Such \textit{continual learning} pose severely challenge for standard DNNs, where previous experiences easily get overwritten as more data and new tasks arrive, i.e., the notorious \textit{catastrophic forgetting}~\citep{goodfellow2013empirical,mccloskey1989catastrophic}. 

\begin{wrapfigure}{r}{0.52\linewidth}
\vspace{0.4em}
\centering
\includegraphics[width=1\linewidth]{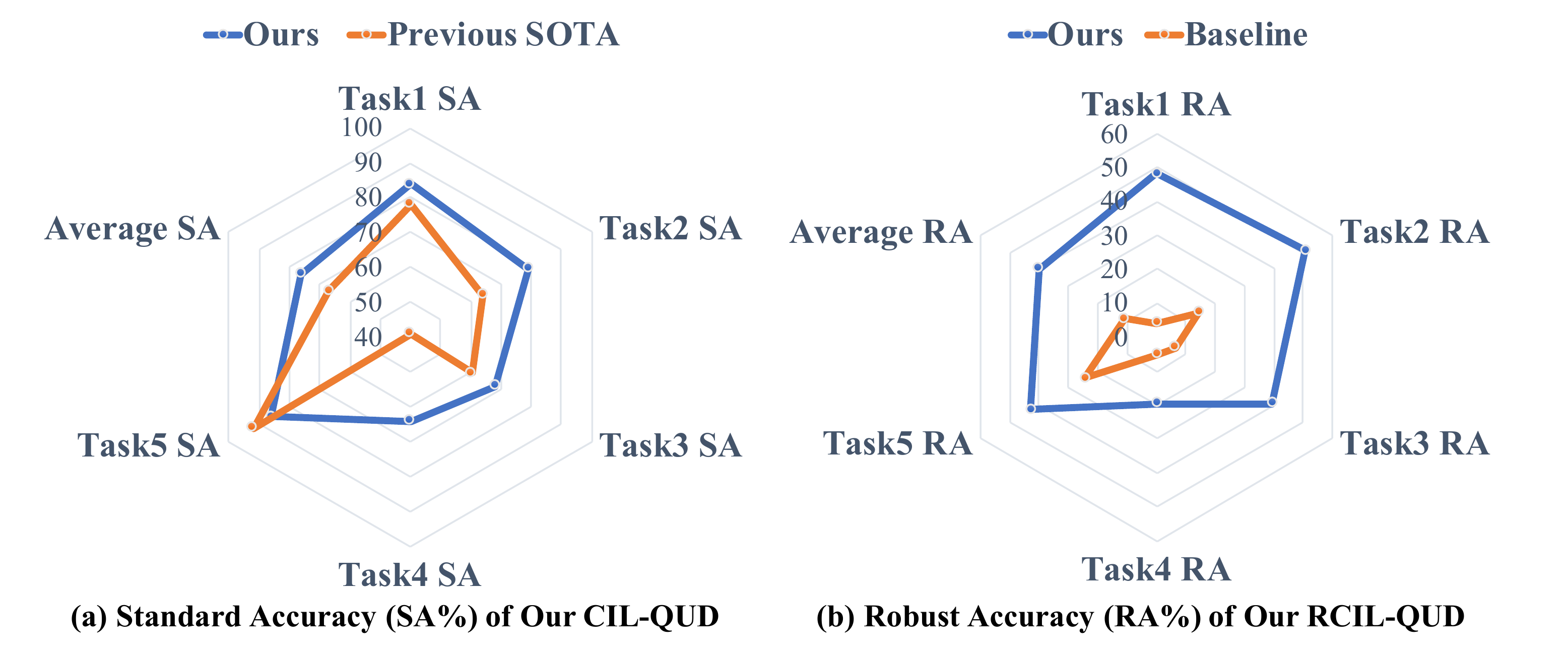}
\vspace{-1.5em}
\caption{\small Summ\RV{a}ry of our achieved performance on CIFAR-10, where CIL is conducted over $5$ incremental learning tasks (each $2$-class). Figure (a) presents the standard accuracy (SA) achieved by CIL-QUD versus the previous SOTA~\citep{Belouadah_2019_ICCV}. Figure (b) shows the robust accuracy (RA) achieved by RCIL-QUD versus the baseline of directly applying adversarial training~\citep{madry2017towards} to CIL.}
\vspace{-0.6em}
\label{fig:tisser}
\end{wrapfigure}

This paper is focused on a realistic yet challenging setting of continual learning, called \textit{class-incremental learning} (CIL)~\citep{rebuffi2017icarl,Belouadah_2019_ICCV,belouadah2020scail,zhang2019class}. In CIL,  the classifier model will need to be incrementally re-trained from time to time, when new classes are added. Ideally, the re-training should provide a competitive multi-class classifier for all classes observed so far at any time. Unfortunately, naively augmenting and fine-tuning the model to learn new classes will only see an abrupt degradation of performance on the original set of classes, when the training objective is adapted to the newly added set of classes. Several attempts have been made by storing past training data~\citep{castro2018end,javed2018revisiting,rebuffi2017icarl,Belouadah_2019_ICCV,belouadah2020scail}, by generative models~\citep{kemker2017fearnet,shin2017continual,li2020incremental}, or by regularized fine-tuning~\citep{aljundi2017expert,li2017learning}. However, they suffer from either excessive memory overhead or the so-called asymmetric information problem~\citep{zhang2019class}, learning this CIL problem far from being resolved. More detailed discussions will be presented in Section 2.

We instead embrace another potential blessing: using publicly available unlabeled auxiliary data in CIL, which can be obtained at negligible costs (e.g., crawled from the web). The power of unlabeled data has recently drawn explosive interests, in multiple contexts such as semi-supervised learning~\citep{chen2020big}, self-training~\citep{xie2020self}, and self-supervised learning~\citep{chen2020improved,chen2020simple}. With massive ``free'' unlabeled data serving pre-training or regularization, those prior works significantly reduced their reliance on labeled data, sometimes performing on par with fully supervised classification while using one to two magnitudes less labeled samples.  

Particularly for CIL, the training data and labels are not only expensive to \textit{collect} (same as standard learning), but also expensive to \textit{store} after its own training task finishes. The memory cost can become predominant, and further in many cases the legacy data cannot remain accessible after training, due to legal or proprietary reasons. That naturally makes external unlabeled data even more promising for CIL: it is a cheap substitute for past training data and needs not to be always stored. Lately, \citet{zhang2019class} presented a new paradigm that proved the concept of improving CIL with external unlabeled data. Each time as new tasks arrive, the authors first trained a separate new model for the new classes using labeled data, and then distilled the two teacher models (the old and new models) into one student model. The second step leveraged unlabeled auxiliary data for distillation, which the authors claimed would help debias the knowledge transferred from both teacher models, compared to using either old or new training data. Their method delivered solid accuracies on CIL classification benchmarks. However, their proposed method was limited by the inefficiency of performing tedious two-step training each time (first the new teacher model, and then the distillation); and the distillation step uses all unlabeled data ($\sim$1 million images in their default case), which further causes significant training burdens.

\subsection{Our Contributions}

We seek to push forward the utility and potential of unlabeled data in CIL, by asking two further questions:\vspace{-0.3em}
\begin{flushleft}
      \textit{\textbf{Q1:}Provided with massive unlabeled data, can we sample and leverage them with higher efficiency?} 
      \\
    \textit{\textbf{Q2:} From unlabeled data, can we harvest more ``bonus'' in other CIL performance dimensions besides accuracy?}
\end{flushleft}

\underline{Our answer to \textbf{Q1}} draws the best ideas from two worlds. On the one hand, existing methods that store and repl\RV{a}y past training data~\citep{castro2018end,javed2018revisiting,rebuffi2017icarl,Belouadah_2019_ICCV,belouadah2020scail} are still the most effective in overcoming catastrophic forgetting, despite its storage headache and sometimes the data privacy/copyright concerns. On the other hand, \citet{zhang2019class} pioneered on CIL with unlabeled data, but sacrificed training/data efficiency as above explained. To mitigate their respective challenges, we propose to integrate the complementary strengths of \textit{storing past data} and \textit{leveraging unlabeled data}; specifically, we only store a handful of historical samples, which would be used as \textit{anchor points} to \textit{query the most relevant unlabeled data}. Aided by the learning-without-forgetting (LwF) regularizer~\citep{li2017learning}, those queried unlabeled samples join the labeled data from the newly added class to balance between preserving the historical and learning the new knowledge. Our ablation experiments endorse that (1) such queried unlabeled data leads to state-of-the-art (SOTA) CIL performance while sacrificing no efficiency; and (2) using anchor-based query stably outperforms random selection, and stays robust to the unlabeled data distribution shifts and volume variations.

\underline{Our answer to \textbf{Q2}} is strongly motivated by the prevailing success of utilizing unlabeled data besides standard classification, e.g., improving robustness~\citep{alayrac2019labels}. We hope to extend and validate those benefits to CIL as well. Specifically, adversarial robustness~\citep{chen2020adversarial} arises as a key demand when deploying DNNs to safety/security-critical applications. While continual learning has been so far focused on maintaining \textit{accuracy} across the stream of tasks, we consider it the same necessary - if not more - to examine whether the model can maintain \textit{robustness} across old and new tasks: a critical step towards enabling trustworthy learning in the open world. This problem has been unfortunately largely overlooked and under-explored in the CIL regime. In fact, even one-shot transferability~\citep{pmlr-v97-hendrycks19a,shafahi2019adversarially,goldblum2019adversarially,chan2019thinks} of robustness (from one source domain pre-trained model, to one target domain new task) has not been studied until recently, and was shown to be challenging. We fill in this research gap, by (1) for the first time, studying the \textit{catastrophic forgetting of robustness}\footnote{It is similar to the catastrophic forgetting of standard generalization but with different evaluation metrics. In other words, robustifying models on a new task leads to robustness degradations on previous tasks, as shown in Figure~\ref{fig:tisser} (\textit{right}).} and showing that it cannot be trivially fixed; and (2) extending our new framework with unlabeled data query to sustaining strong robustness in the CIL scheme, with several robustified regularizations. That provides both an extra benchmark dimension for CIL, and a significant advance in the study of transferable robustness beyond one-shot. 

Our specific contributions are summarized below: 
\vspace{-2mm}
\begin{itemize}
    \item A novel framework of \textit{Class-Incremental Learning with Queried Unlabeled Data} (\textbf{CIL-QUD}), that seamlessly integrates a handful of stored historical samples and unlabeled data through anchor-based query. CIL-QUD also carefully leveraged LwF regularizers~\citep{li2017learning} and balanced training~\citep{zhang2019balance} as building blocks. It has light overheads in the model, data storage, and training.
    \item A first-of-its-kind study of preserving adversarial robustness in CIL, and an extension of CIL-QUD to its robustified version called \textbf{RCIL-QUD}. Specifically, we propose and compare two robust versions of LwF regularizers together with an add-on robust regularizer, which are built on the queried unlabeled data.
    \item Experiments demonstrating that the power of unlabeled data can effectively extend to CIL, contributing substantially to superior accuracy as well as adversarial robustness - both preserved without catastrophic forgetting. In Figure~\ref{fig:tisser}, CIL-QUD outperforms previous SOTA by large margins of $10.28\%$ accuracy on CIFAR-10; on CIFAR-100, it also outperforms~\citep{zhang2019class} by a $1.19\%$ margin. RCIL-QUD further establishes the first strong milestone for robustness-aware CIL, that significantly surpasses existing baselines.
\end{itemize}

\section{Related Work}
\vspace{-2mm}
\paragraph{Class-incremental Learning} Among numerous methods developed, one category of approaches~\citep{wang2017growing,rosenfeld2018incremental,rusu2016progressive,aljundi2017expert,rebuffi2018efficient,mallya2018piggyback} incrementally grow the model capacity to accommodate new classes, yet suffering from the explosive model size as well as inflexibility (e.g., requiring task ID at inference). Another category of solutions is based on transfer learning~\citep{kemker2017fearnet,belouadah2018deesil}, whose effectiveness yet hinges on the pre-training quality. 

A popular and successful family of CIL methods~\citep{li2017learning,castro2018end,javed2018revisiting,rebuffi2017icarl,Belouadah_2019_ICCV,belouadah2020scail} (partially) memorized past training data to fight catastrophic forgetting, bypassing the need for dynamic model capacity. Many of those algorithms viewed CIL as an imbalanced learning problem, where previous and newly added classes made an class size extreme disparity~\citep{he2009learning,buda2018systematic}. Learning without Forgetting (\textit{LwF})~\citep{li2017learning} made an attempt to fix that dilemma through an LwF regularization via knowledge distillation. Although the original \textit{LwF} did not deposit past data, later works~\citep{rebuffi2017icarl} augmented it with a memory bank of previous tasks and previously stored data. More follow-ups~\citep{castro2018end,he2018exemplar,javed2018revisiting,rebuffi2017icarl,Belouadah_2019_ICCV,belouadah2020scail} advanced the LwF regularization further. For example, \textit{IL2M}~\citep{Belouadah_2019_ICCV} utilized a second memory bank to store past class statistics obtained at past training, and incorporated those stored statistics to compensate for the previous (minority) classes' predictions. 

Since storing past data inevitably incurred memory overheads, \citet{he2018exemplar} used GANs to generate exemplars for previous tasks, for supplying a re-balanced training set at each incremental state. One of the most relevant works to ours is~\citet{zhang2019class}, which pioneered the usage of free external unlabeled data to boost CIL, as we have discussed in Section 1. In comparison, \textbf{our proposed  CIL-QUD clearly distinguishes itself} from~\citet{zhang2019class}, by (1) using adaptively queried unlabeled samples based on stored anchors, instead of full unlabeled data; (2) avoiding two-stage training and instead adopting LwF-regularized fine-tuning; (3) combining training with class-balanced and randomly sampled data in CIL~\citep{zhang2019balance}. Besides, neither \citet{zhang2019class} nor any other mentioned work ever touched the robustness preservation in CIL. More details on those differences are expanded in Section~\ref{sec:methods}. The other relevant study of leveraging unlabeled data in CIL is~\citet{lee2019overcoming}, which designs a confidence-based sampling and a new distillation approach, to integrate external free knowledge and mitigate the catastrophic forgetting. Compared to \citet{lee2019overcoming}, we get rid of the multi-stage training, investigate diverse LwF regularizers, and introduce an effective dual-classifier backbone with a novel proposed ensemble mechanism.

On the other hand, recent investigations~\citep{wang2021ordisco,he2021unsupervised} also explore other alternative possibilities of leveraging unlabelled data to boost continual learning. For example, \citet{wang2021ordisco} replays unlabeled data sampled from a conditional generator, and utilizes a consistency regularization to learn an improved continual classifier. \citet{he2021unsupervised} studies continual learning in a fully unsupervised mode by assigning unlabeled data with clustered pseudo labels. Meanwhile, \citet{bateni2022beyond} and \citet{chen2020long} enable few-shot and efficient continual learning respectively, with the assistance of unlabeled data. A recent survey paper~\citep{qu2021recent} also provides a good summary of current achievements in this field.

\vspace{-0.6em}
\paragraph{Adversarial Robustness and Its Transferablity} 
DNNs commonly suffer from adversarial vulnerability~\citep{goodfellow2014explaining}, and numerous defense methods have been invented~\citep{madry2017towards,sinha2017certifiable,rony2019decoupling,zhang2019theoretically,ding2020mma}. However, most of them focus on attacking/defending DNNs trained on a single fixed dataset and task. Recently, a couple of works emerge to cogitate the transferability of adversarial robustness~\citep{pmlr-v97-hendrycks19a,shafahi2019adversarially,goldblum2019adversarially,chan2019thinks,chen2020adversarial}, from a robust model pre-trained on a source domain to another target domain. They revealed that directly fine-tuning on the target domain data will quickly overwrite the pre-trained robustness~\citep{chen2020adversarial}. One needs to refer to either adversarial fine-tuning on the target domain~\citep{chen2020adversarial}, or specific regularizations such as knowledge distillation or gradient matching to preserve the source domain robustness knowledge~\citep{shafahi2019adversarially,goldblum2019adversarially,chan2019thinks}. In comparison, we target at inheriting robustness across many sets of new data and tasks added continually, where the conventional transfer learning could be viewed as its oversimplified case. To our best knowledge, \textbf{our proposed RCIL-QUD marks the first-ever effort to explore this new daunting setting}.

\section{Class-Incremental Learning with Queried Unlabeled Data (CIL-QUD)} \label{sec:methods}

In this section, we begin by presenting a brief background on class-incremental learning (CIL), and show our CIL-QUD that overcomes catastrophic forgetting by adaptively querying and leveraging unlabeled data. 

\subsection{CIL Preliminaries and Setups}

\begin{wrapfigure}{r}{0.52\linewidth}
    \centering
    \vspace{-4mm}
    \includegraphics[width=0.90\linewidth]{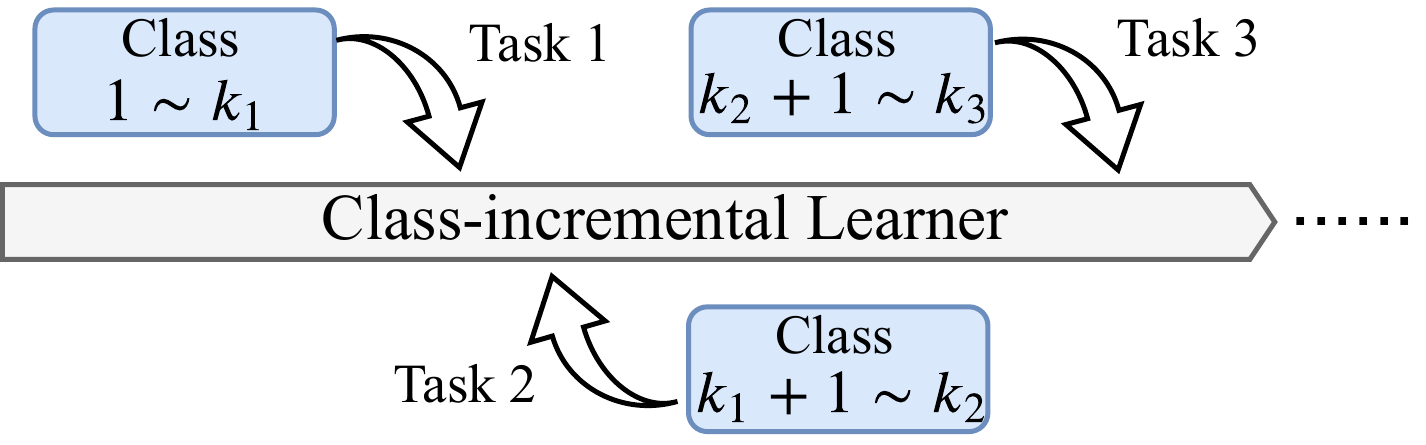}
    \vspace{-3mm}
    \caption{\small CIL learns continuously from a sequential data stream where new tasks are added. Each new task $i$ may contain several classes from class $k_{i-1}+1$ to $k_i$. The CIL learner aims for multi-class classification for both previous and newly added classes.}
    \label{fig:CIL}
    \vspace{-4mm}
\end{wrapfigure}

As in Figure~\ref{fig:CIL}, CIL models are continuously trained over a \textit{sequential data stream}, where a new classification task (consisting of \textit{unseen classes}) could be added every time. This makes CIL highly challenging in contrast to static learning and conventional (one-shot) transfer learning.
Following~\citet{castro2018end,he2018exemplar,rebuffi2017icarl}, we consider a practical and challenging CIL setting with only a small memory bank $\mathcal{S}$ to store data from previous classes. Neither task ID nor order is pre-assumed to be known.

Let $\mathcal{T}_1,\mathcal{T}_2,\cdots$ represent a sequence of CIL tasks, and the $i$th task $\mathcal T_i$ contains data that fall   in $(k_i - k_{i-1})$ classes 
$\mathcal{C}_i=\{c_{k_{i-1}+1},c_{k_{i-1}+2},\cdots,c_{k_i}\}$, with $k_0 = 1$ by convention.
At the $i_{\mathrm{th}}$ incremental learning session, we only have access to the training data associated with $\mathcal{T}_i$, and a limited number of stored data in $\mathcal{S}$ from the previous learning tasks. Let $f(\boldsymbol{\theta},\boldsymbol{\theta}_\mathrm{c}^{(i)},\mathbf{x})$ denotes the mapping from input samples $\mathbf x\in\bigcup_{j=1}^{i}\mathcal{T}_j$ to the corresponding classes $\bigcup_{j=1}^{i}\mathcal{C}_j$  acquired from $\mathcal{T}_1,\cdots,\mathcal{T}_i$. Here $(\boldsymbol \theta, \boldsymbol \theta_\mathrm{c}^{(i)})$ denotes the parameters of a CIL model updated till task $\mathcal T_i$. To be more specific, the CIL model consists of a feature extractor $\varphi(\boldsymbol{\theta}, \mathbf x) \in \mathbb R^d$ with parameters  $\boldsymbol{\theta}$, which maps the input $\mathbf x \in \bigcup_{j=1}^{i}\mathcal{T}_j $ to a 
$d$-dimensional feature space. It is then followed by a  multi-class (i.e., $\bigcup_{j=1}^{i}\mathcal{C}_j$) classifier
with parameters $\boldsymbol{\theta}_{\mathrm c}^{(i)}$, which maps the $d$-dimensional feature to the prediction vector $\rho_i(\boldsymbol{\theta},\boldsymbol{\theta}_{\mathrm{c}}^{(i)}, \mathbf x)$ at the current task $\mathcal{T}_i$.

\vspace{-3mm}
\paragraph{Problem statement.} We now formally define our target CIL problem: Given a previously trained model $f(\hat{\boldsymbol{\theta}},\hat{\boldsymbol{\theta}}_\mathrm{c}^{(i-1)},\mathbf x)$ under tasks $\mathcal{T}_1,\cdots,\mathcal{T}_{i-1}$, the objective is to obtain an updated model $f(\boldsymbol{\theta},\boldsymbol{\theta}_\mathrm{c}^{(i)},\mathbf x)$ to preserve the generalization ability and robustness among all learned tasks, even if one can only have the access to data from the memory bank $\mathcal{S}$ as well as the current task $\mathcal{T}_i$. Here the feature extractor $\varphi(\boldsymbol{\theta},\mathbf x)$ is shared over all seen tasks, but the scale of the multi-class classifier linearly increases along with the incremental classes.

\subsection{Anchor-based Query of Unlabeled Data} 
Catastrophic forgetting poses the major challenge to CIL. A major reason of forgetting is the asymmetric information between previous classes and newly added classes at each incremental learning stage. Existing CIL approaches~\citep{he2009learning,buda2018systematic,rebuffi2017icarl,Belouadah_2019_ICCV,belouadah2020scail,chu2016best,kemker2017fearnet,parisi2019continual} undertake the forgetting dilemma by training on new task data plus stored previous data. However, the inevitable storage limitation could still cause severe prediction biases as more classes come in. \citet{chu2016best,kemker2017fearnet,parisi2019continual} introduced balanced fine-tuning, yet incurring the risk of over-fitting new/minority classes. \citet{zhang2019class} tried to fix the dilemma by referring external unlabeled data. They did not store any past training data; but as above discussed suffered from considerable training overhead.

We present our novel remedy of utilizing anchor-based unlabeled data query, to balance between preserving the historical and learning the new knowledge. \underline{First}, we store a small number of i.i.d randomly picked samples as ``anchors” for every past training class, e.g., as few as ten samples per class on CIFAR-100. \underline{Then}, during the next incremental processes, we query more auxiliary unlabeled samples from public sources with stored anchors, using certain similarity matches (see Section 3.4 for details). The queried samples are expected to present ``similar'' and more relevant information to past training data, compared to random samples. Practically, unlabeled samples can be queried from public sources containing diverse enough natural images, e.g., Google Images, that are not necessarily tied with previous classes. \underline{Next}, we inject the previous information into learning new classes, by tuning with the learning-without-forgetting (LwF) regularizers~\citep{li2017learning} on queried unlabeled data.  

We next detail on the concrete regularizers $\mathcal{L}_{\mathrm{LwF}}$ used. Let $\mathcal{U}$ donates the queried unlabeled data, $\mathcal{L}_{\mathrm{LwF}}$ can be chosen from either knowledge distillation ($\mathcal{KD}$\footnote{$\mathcal{KD}$ here denotes the regularization function of knowledge distillation, a modified cross-entropy as in~\citet{li2017learning}; $\mathcal{FT}$ denotes the regularization function of feature transferring, an $\ell_p$ distance metric as in~\citet{shafahi2019adversarially}.})~\citep{hinton2015distilling,li2017learning} or feature transferring ($\mathcal{FT}^{\textcolor{red}{1}}$)~\citep{shafahi2019adversarially}, i.e., $\mathcal{L}_{\mathrm{LwF}}\in\{\mathcal{KD},\mathcal{FT}\}$.

$\mathcal{KD}$ is one of the most classical regularizers in CIL~\citep{li2017learning,castro2018end,he2018exemplar,javed2018revisiting,rebuffi2017icarl,Belouadah_2019_ICCV,belouadah2020scail}, but was usually applied on fully-stored previous data and/or newly added data. In our case, we enforce the output probabilities $\rho(\boldsymbol{\theta},\boldsymbol{\theta}_\mathrm{c},\mathbf x)$ of each queried unlabeled image $\mathbf x \in \mathcal{U}$ to be close to the recorded $\rho(\hat{\boldsymbol{\theta}},\hat{\boldsymbol{\theta}}_\mathrm{c},\mathbf x)$. The $\mathcal{KD}$ regularization is then given by:
{\small\begin{align}
 \hspace*{-0.15in}   \displaystyle \mathcal L_{\mathrm{LwF}}(\boldsymbol{\theta}, \boldsymbol{\theta}_{\mathrm{c}}) \Def  \mathbb E_{\mathbf x \in \mathcal{U}} \left[\mathcal{KD} \left(\rho(\boldsymbol{\theta}, \boldsymbol{\theta}_{\mathrm{c}},\mathbf x),\rho(\hat{\boldsymbol{\theta}}, \hat{\boldsymbol{\theta}}_{\mathrm{c}},\mathbf x)\right) \right] \hspace*{-0.15in}   
    \label{eq:kd}
\end{align}}%
where $\boldsymbol{\theta}$ and $\boldsymbol{\theta}_{\mathrm{c}}$ present the parameters of current feature extractor and classifiers respectively, and $\hat{\boldsymbol{\theta}}$ and $\hat{\boldsymbol{\theta}}_\mathrm{c}$ stand for the previous ones. 

$\mathcal{FT}$~\citep{shafahi2019adversarially} inherits previous knowledge by maximizing the similarity between current feature representations $\varphi(\boldsymbol{\theta},\mathbf x)$ and previous features $\varphi(\hat{\boldsymbol{\theta}},\mathbf x)$ on unlabeled data $\mathbf x \in \mathcal{U}$:
{\small\begin{align}
    \begin{array}{ll}
     \displaystyle \mathcal L_{\mathrm{LwF}}(\boldsymbol{\theta}) \Def  \displaystyle \mathbb E_{\mathbf x \in \mathcal{U}}\left[\mathcal{FT}(\varphi(\boldsymbol{\theta},\mathbf x),\varphi(\hat{\boldsymbol{\theta}},\mathbf x))\right],
    \end{array} \label{eq:feature}
\end{align}}%
where $\varphi(\cdot)$ is the feature extractor, $\boldsymbol{\theta}$ is defined the same as above, and $\mathcal{FT}$ is a distance metric, which we choose to be $\ell_1$ norm here. 
An \textit{interesting finding} from our later experiments (Table~\ref{table:standard} and~\ref{table:adversarial_RA}) is that, $\mathcal{KD}$ outperforms $\mathcal{FT}$ in preserving generalization ability under standard CIL, but $\mathcal{FT}$ becomes more useful in the later robustness-aware CIL.  

\subsection{Balanced Training with Auxiliary Classifiers} \label{balance_training}
\begin{figure}[t]
    \centering
    \includegraphics[width=1\linewidth]{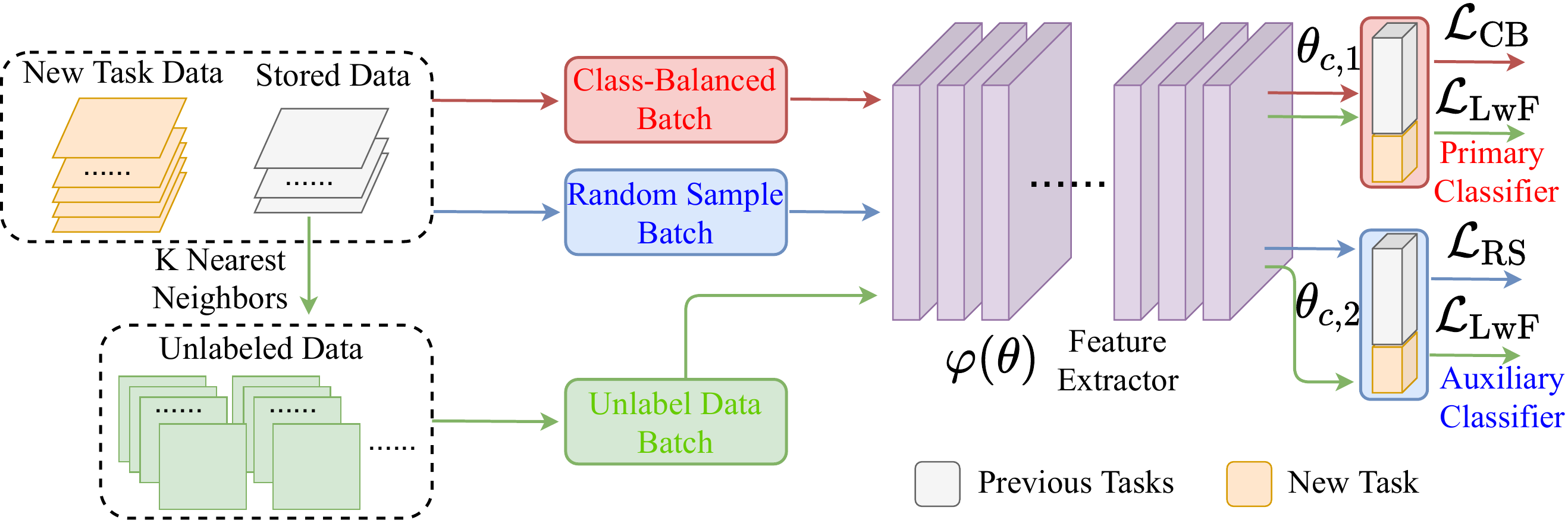}
    \vspace{-7mm}
    \caption{\small Overall framework of CIL-QUD. From left to right, before each training iteration, we first create three batches: \textcolor{red}{Class-Balanced Batch} $\mathcal{B}_{\mathrm{CB}}$ contains balanced data between stored previous classes and newly added classes; \textcolor{blue}{Random Sample Batch} $\mathcal{B}_{\mathrm{RS}}$ includes data randomly sampled from all stored data and new task data; \textcolor{green}{Unlabeled Data Batch} $\mathcal{B}_{\mathrm{UD}}$ consists of auxiliary unlabeled data queried by stored previous data using the K-nearest-neighbors algorithm over the feature embeddings. Then, three batches are fed into a feature extractor $\varphi(\cdot)$, ResNet-18~\citep{he2016deep}. \textit{Red arrows} (\textcolor{red}{\ding{212}}), \textit{Blue arrows} (\textcolor{blue}{\ding{212}}) and \textit{Green arrows} (\textcolor{green}{\ding{212}}) represent the corresponding feed forward paths. The primary classifier takes features from $\mathcal{B}_{\mathrm{CB}}$ and $\mathcal{B}_{\mathrm{UD}}$ to calculate the objective $\mathcal{L}_{\mathrm{CB}}$ and $\mathcal{L}_{\mathrm{LwF}}$. The auxiliary classifier produces the objective $\mathcal{L}_{\mathrm{RS}}$ and $\mathcal{L}_{\mathrm{LwF}}$ with features from $\mathcal{B}_{\mathrm{RS}}$ and $\mathcal{B}_{\mathrm{UD}}$. Please zoom-in for details.}
    \vspace{-3mm}
    \vspace{-0.5em}
    \label{fig:CBS}
\end{figure}

Another well-known remedy for the imbalanced classes is to re-sample mini-batches~\citep{chawla2002smote,haixiang2017learning,tahir2009multiple,wang2019dynamic,zhang2019balance}. It artificially re-balances between majority classes and minority classes to alleviate the negative effect of skewed training data distributions. In this paper, spurred by the recent progress in long-tail visual recognition~\citep{zhang2019balance,lee2019overcoming}, we for the first time combine both \textit{random and class-balanced sampling} strategies with auxiliary classifiers into CIL.  Our designed network architecture $f(\boldsymbol{\theta},\boldsymbol{\theta}_\mathrm{c},\mathbf x)$ for CIL-QUD contains a \textit{shared feature extractor} $\varphi(\boldsymbol{\theta},\mathbf x)$, a \textit{primary classifier} trained with class-balanced data, and an \textit{auxiliary classifier} trained with randomly sampled data~\citep{zhang2019balance}; see  Fig.~\ref{fig:CBS} and Sec.\,\ref{sec: overall_framework} for detailed illustration. In this way, the feature extractor learns from both class-balanced and randomly sampled data, where the former prevents the feature extractor from prediction preference toward majority classes, and the latter improves generalization in minority classes.

We notice that both classifiers see satisfying results, but there also exists a performance trade-off~\citep{zhang2019balance} between them. The primary classifier, trained with class-balanced data, remembers more previous knowledge and performs better on the previous/minority classes; the auxiliary classifier, trained with randomly sampled data, performs better on the new/majority classes, at the cost of forgetting previous knowledge to some extent. That trade-off motivates us to design a \textbf{c}lassifier \textbf{e}nsemble \textbf{m}echanism (CEM) for extra performance boosts. More details are referred to the supplement Section~\ref{sec:SEM} and Algorithm~\ref{alg:vote}.

\subsection{Overall Framework of CIL-QUD}
\label{sec: overall_framework}
Our proposed framework, CIL-QUD, effectively integrates the above two remedies, as presented in Figure~\ref{fig:CBS}. \underline{First}, we create Class-Balanced Batch ($\mathcal{B}_{\mathrm{CB}}$) and Random Sample Batch ($\mathcal{B}_{\mathrm{RS}}$) through class-balanced and random sampling from stored previous data and incoming new data. $\mathcal{B}_{\mathrm{CB}}$ and $\mathcal{B}_{\mathrm{RS}}$ are both used in feature extractor training, while they are separately employed to the primary and the auxiliary classifier. \underline{Second}, we query the auxiliary unlabeled data by finding K-nearest-neighbors of previously stored labeled data over their feature embeddings. Details on the use of unlabeled data and feature embeddings can be found in section~\ref{sec:setup}. We then randomly sample from queried unlabeled data to form an Unlabeled Data Batch ($\mathcal{B}_{\mathrm{UD}}$). In this way, we inject more ``similar'' knowledge of previous classes into CIL models via $\mathcal{B}_{\mathrm{UD}}$. \underline{Third}, feeding the $\mathcal{B}_{\mathrm{UD}}$ together with $\mathcal{B}_{\mathrm{CB}}$ and $\mathcal{B}_{\mathrm{RS}}$, the primary classifier produces a cross-entropy loss $\mathcal{L}_{\mathrm{CB}}$ for classification and a regularization term $\mathcal{L}_{\mathrm{LwF}}$ for preventing forgetting. Similarly, the auxiliary classifier yields $\mathcal{L}_{\mathrm{RS}}$ and $\mathcal{L}_{\mathrm{LwF}}$. Note that $\mathcal{L}_{\mathrm{LwF}}$ is calculated on the unlabeled data batch, i.e, $\mathcal{B}_{\mathrm{UD}}$. In brief, CIL-QUD is cast as the following regularized optimization problem:
\begin{align}
    \begin{array}{ll}
    \displaystyle \min_{\boldsymbol{\theta},\boldsymbol{\theta}_{\mathrm{c},1}, \boldsymbol{\theta}_{\mathrm{c},2}}  \mathbb E_{(\mathbf x, y) \in \mathcal{B}_{\mathrm{CB}}} \left[\mathcal{L}_{\mathrm{CB}}(f(\boldsymbol{\theta}, \boldsymbol{\theta}_{\mathrm{c},1},\mathbf x), y)\right] & + \mathbb E_{(\mathbf x, y) \in \mathcal{B}_{\mathrm{RS}}} \left[\mathcal{L}_{\mathrm{RS}}(f(\boldsymbol{\theta},\boldsymbol{\theta}_{\mathrm{c},2}, \mathbf x), y) \right] \\ & + 
    \lambda \cdot \left[\mathcal{L}_{\mathrm{LwF}}(\boldsymbol{\theta}, \boldsymbol{\theta}_{\mathrm{c},1}) +  \mathcal{L}_{\mathrm{LwF}}(\boldsymbol{\theta}, \boldsymbol{\theta}_{\mathrm{c},2}) \right],
    \end{array} \label{eq:standard}
\end{align}
where $\boldsymbol{\theta}_{c,1},\boldsymbol{\theta}_{c,2}$ donate the parameters from the primary and auxiliary classifiers respectively, $\lambda$ is a hyperparameter, controlling the contributions of LwF regularizers on queried unlabeled data. In our case, $\lambda=0.5$. For tuning of hyper parameters, we perform a grid search. $\mathcal{L}_{\mathrm{LwF}}$ is selected from $\{\mathcal{KD},\mathcal{FT}\}$ depicted in Equation~\ref{eq:kd} and~\ref{eq:feature}. The auxiliary classifier serves as an implicit regularizer for preventing networks from over-fitting minority classes of previous data.

\vspace{-0.5em}
\section{Robustified Class-Incremental Learning with Queried Unlabeled Data (RCIL-QUD)} 
In this section, we motivate the new setup of \textit{robustness-aware CIL}, where a model has to maintain adversarial robustness across old and new tasks. Assisted by our unlabeled data query scheme, we propose two robustified LwF regularizers together with an add-on robust regularizer that can be plugged in CIL-QUD, leading to the RCIL-QUD.

The vulnerability of DNNs raises critical demands for improving their robustness~\citep{goodfellow2014explaining}. However, no formal assessment of adversarial robustness has been performed in the CIL setting. It is natural to suspect that catastrophic forgetting would make learned robustness hard to sustain over new tasks too. Indeed, recent studies have identified such challenges, even in the much simpler case of transferring robustness from one source to another target domain~\citep{pmlr-v97-hendrycks19a,shafahi2019adversarially,goldblum2019adversarially,chan2019thinks}. Moreover, compared to standard generalization, DNNs need significantly more data to achieve adversarially robust generalization~\citep{schmidt2018adversarially,madry2017towards,shafahi2019adversarially,kurakin2016adversarial,zhai2019adversarially}, which also challenges CIL where previous classes may only have a handful of stored samples - \textit{that is precisely \textbf{why} our queried external data can become the necessary aid and the blessing}.

To handle this new daunting setting, we first investigate the \textbf{catastrophic forgetting of robustness}. As shown in Figure~\ref{fig:tisser} (b) and later in Section 5.3, neither the conventional LwF techniques nor the direct application of adversarial training (AT)~\citep{madry2017towards} (the most successful defense method) can prevent the model's learned robustness from decaying over time. Here the direct application of AT refers to train over the worst-case losses penalized by standard LwF regularizers.

In our proposed RCIL-QUD, in addition to incorporating worst-case (min-max) training losses as AT, we propose to robusify our LwF regularizations with queried unlabeled data, denoted as $\mathcal{L}^{\mathrm{R}}_{\mathrm{LwF}}$, for sustaining adversarial robustness in the CIL scenario. RCIL-QUD formulation is depicted as: 
\vspace{-2mm}
{\small\begin{align}
    \begin{array}{ll}
    \displaystyle \min_{\boldsymbol{\theta},\boldsymbol{\theta}_{\mathrm{c},i},\boldsymbol{\theta}_{\mathrm{c},2}} \mathbb E_{(\mathbf x, y) \in \mathcal{B}_{\mathrm{CB}}} \left[\displaystyle \max_{\|\boldsymbol{\delta}\|_{\infty}\le\epsilon}\mathcal{L}_{\mathrm{CB}}(f(\boldsymbol{\theta}, \boldsymbol{\theta}_{\mathrm{c},1},\mathbf x+\boldsymbol{\delta}), y) \right] & +  \displaystyle \mathbb E_{(\mathbf x, y) \in \mathcal{B}_{\mathrm{RS}}}  \left[\displaystyle\max_{\|\boldsymbol{\delta}\|_{\infty}\le\epsilon}\mathcal{L}_{\mathrm{RS}}(f(\boldsymbol{\theta},\boldsymbol{\theta}_{\mathrm{c},2}, \mathbf x+\boldsymbol{\delta}), y) \right]\\& 
    + 
\gamma_1\cdot
\mathcal{L}_{\mathrm{LwF}}^{\mathrm{R}}(\boldsymbol{\theta},\boldsymbol{\theta}_{\mathrm{c}}) + \gamma_2\cdot \mathcal{L}_{\mathcal{RTC}}(\boldsymbol{\theta},\boldsymbol{\theta}_{\mathrm{c}}),
    \end{array} \label{eq:adv}
\end{align}}%
where $\gamma_1$, $\gamma_2$ balance the effect between AT and robust regularizers. In our case, $\gamma_1=0.05, \gamma_2=0.2$. $\delta$ is the adversarial perturbation generated by Projective Gradient Descent (PGD)~\citep{madry2017towards}. $\epsilon$ is the upper bound of perturbations under $\ell_{\infty}$ norm. In practice, RCIL-UD can involve $\mathcal{L}_{\mathrm{LwF}}^{\mathrm{R}}$ implementations from $\{\mathcal{RKD},\mathcal{RFT}\}$ together w. or w.o. $\mathcal{RTC}$, in Equations~\ref{eq:robust_kd},~\ref{eq:robust_feature} and~\ref{eq:trades} below, to pass on historic robustness.

\vspace{-1em}
\paragraph{Robustified knowledge distillation regularizers ($\mathcal{RKD}$).} As standard LwF regularizers are found to be unsuitable for preserving robustness, we thus propose a robustified knowledge distillation term, $\mathcal{RKD}$, via introducing min-max optimization into $\mathcal{KD}$:
{\small\begin{align}
  \hspace*{-0.1in}  \begin{array}{ll}
    \displaystyle \displaystyle \mathbb E_{\mathbf x \in \mathcal{U}} \left[ \max_{\|\boldsymbol{\delta}\|_{\infty}\le\epsilon}\mathcal{KD} \left(\rho(\boldsymbol{\theta}, \boldsymbol{\theta}_{\mathrm{c}},\mathbf x+\boldsymbol{\delta}),\rho(\hat{\boldsymbol{\theta}}, \hat{\boldsymbol{\theta}}_{\mathrm{c}},\mathbf x)\right) \right],
    \end{array}  \hspace*{-0.1in} \label{eq:robust_kd}
\end{align}}%
where $\epsilon$, $\boldsymbol{\delta}$, $\rho$, $\boldsymbol{\theta}$, $\boldsymbol{\theta}_{\mathrm{c}}$, $\boldsymbol{\hat\theta}$ and $\boldsymbol{\hat\theta}_{\mathrm{c}}$ are defined similarly as \eqref{eq:kd} and \eqref{eq:adv}. Our results demonstrate that $\mathcal{RKD}$ is more effective for preserving adversarial robustness on previous classes.

\begin{figure}[t]
    \centering
    \includegraphics[width=0.95\linewidth]{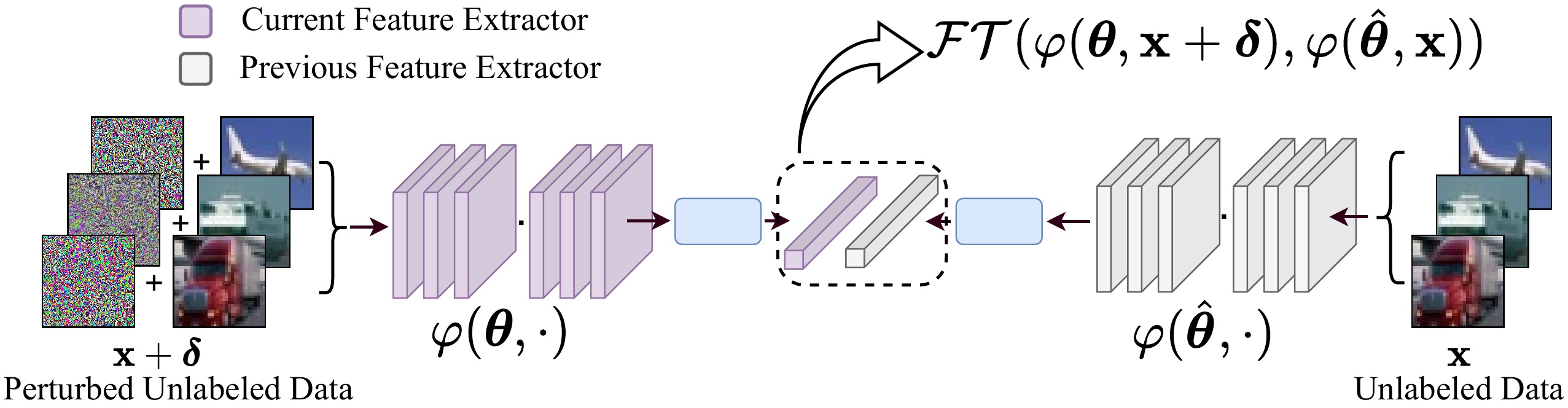}
    \vspace{-3mm}
    \caption{\small Illustrations of $\mathcal{RFT}$ regularization. From left to right, we first feed the perturbed unlabeled data $\mathbf x+\boldsymbol{\delta}$ through current feature extractor $\varphi(\boldsymbol{\theta},\cdot)$ and get a feature vector $\varphi(\boldsymbol{\theta},\mathbf x+\boldsymbol{\delta})$. Then we pass the clean unlabeled data $x$ to previous feature extractor $\varphi(\hat{\boldsymbol{\theta}},\cdot)$ and obtain another feature vector $\varphi(\hat{\boldsymbol{\theta}},\mathbf x)$. $\mathcal{RFT}$ enforces the similarity between two features, where $\mathcal{FT}$ is an $\ell_1$ norm.}
    \label{fig:feature}
    \vspace{-5mm}
\end{figure}

\vspace{-1em}
\paragraph{Robustified feature transferring regularizers ($\mathcal{RFT}$).} We then study the robust version of feature transferring regularizer ($\mathcal{FT}$)~\citep{shafahi2019adversarially}. As mentioned recently in~\citet{shafahi2019adversarially} on robust transfer learning, the feature activations make a main source of robustness. Yet in CIL, the effectiveness of $\mathcal{FT}$ decays considerably as more tasks arrive. The limitation of $\mathcal{FT}$ motivates us to build its novel robustified version, $\mathcal{RFT}$. It enforces the similarity between the current feature representation $\varphi(\boldsymbol{\theta},\mathbf x+\boldsymbol{\delta})$ of perturbed unlabeled data, and the previous representation $\varphi(\hat{\boldsymbol{\theta}},\mathbf x)$ of the clean unlabeled data, as shown in Figure~\ref{fig:feature}. $\mathcal{RFT}$ is mathematically described as:
{\small\begin{align}
    \begin{array}{ll}
    \displaystyle \mathbb E_{\mathbf x \in \mathcal{U}}\left[\max_{\|\boldsymbol{\delta}\|_{\infty}\le\epsilon}\mathcal{FT}(\varphi(\boldsymbol{\theta},\mathbf x + \boldsymbol{\delta)},\varphi(\hat{\boldsymbol{\theta}},\mathbf x))\right],
    \end{array} \label{eq:robust_feature}
\end{align}}%
where $\mathcal{FT}$ has been defined in \eqref{eq:feature}.

\vspace{-2mm}
\paragraph{Robustified TRADES regularizers on CIL ($\mathcal{RTC}$).} 
To further unleash the power of unlabeled data in promoting robustness, we incorporate a SOTA \textbf{r}obust defense called \textbf{T}RADES~\citep{zhang2019theoretically}, as an add-on regularizer over $\mathcal{RKD}$ and $\mathcal{RFT}$. That is owing to the fact that TRADES does not rely on data labels, unlike the adversarial training loss~\citep{madry2017towards}. The formulation is: 
{\small\begin{align}
    \begin{array}{ll}
    \displaystyle \mathbb E_{\mathbf x \in \mathcal{U}} \left[ \displaystyle \max_{\|\boldsymbol{\delta}\|_{\infty}\le\epsilon}\mathcal{KL}\left(\rho(\boldsymbol{\theta}, \boldsymbol{\theta}_{\mathrm{c}},\mathbf x+\boldsymbol{\delta}),\rho(\boldsymbol{\theta}, \boldsymbol{\theta}_{\mathrm{c}},\mathbf x)\right) \right],
    \end{array} \label{eq:trades}
\end{align}}%
where $\rho$ is defined as above, and $\mathcal{KL}$ is Kullback–Leibler divergence. $\mathcal{RTC}$ boosts robustness via minimizing the ``difference'' between the predication probabilities $\rho(\boldsymbol{\theta}, \boldsymbol{\theta}_{\mathrm{c}}, \mathbf x)$ of the current model on clean samples and $\rho(\boldsymbol{\theta}, \boldsymbol{\theta}_{\mathrm{c}}, \mathbf x+\boldsymbol{\delta})$ of the current model on adversarial samples. It is worth mentioning that $\mathcal{RTC}$ \eqref{eq:trades} is different from  $\mathcal{RKD}$ \eqref{eq:robust_kd}, the former enforces the stability of predictions of the same (current) model $(\boldsymbol \theta, \boldsymbol \theta_{\mathrm{c}})$ before and after input perturbations, while the latter promotes the prediction stability of the current model at input perturbations with respect to a reference model, namely, the previously learned  $(\hat{\boldsymbol \theta}, \hat{\boldsymbol \theta}_{\mathrm{c}})$.

\section{Experiments and Analyses}
\vspace{-2mm}
\subsection{Implementation Details}
\paragraph{Dataset, Memory Bank, and External Source.} We evaluate our proposed method on CIFAR-10 and CIFAR-100 datasets~\citep{krizhevsky2009learning}. We randomly split the original training dataset into training and validation set with a ratio of $9:1$.  We use all default data augmentations provided by~\citet{li2017learning}, and image pixels are normalized to $[0,1]$. On CIFAR-10, we divide the $10$ classes into splits of $2$ classes with a random order ($10/2=5$ tasks); On CIFAR-100, we divide $100$ classes into splits of $20$ classes with a random order ($100/20=5$ tasks). Namely, at each incremental time, the classifier dimension will increase by $2$ for CIFAR-10 and $20$ for CIFAR-100. 

We set the memory bank to store $100$ images per class for CIFAR-10 and $10$ images for CIFAR-100 by default. The default external source of queried unlabeled data is 80 Million Tiny Image dataset~\citep{torralba200880}, while another source of ImageNet $32\times32$~\citep{deng2009imagenet} is investigated later. During each incremental session, we query $5,000$ and $500$ unlabeled images per class for CIFAR-10 and CIFAR-100, respectively by default. Increasing the amounts of queried unlabeled data may improve performance further but incur more training costs. We use a buffer of fixed capacity to store $128$ queried unlabeled images at each training iteration. More training and model selection details are referred to the supplement.

\vspace{-1em}
\paragraph{Evaluation Metrics.} We evaluate on ResNet18~\citep{he2016deep} in terms of: (1) Standard Accuracy (SA): classification testing accuracy on the clean test dataset; and (2) Robust Accuracy (RA): classification testing accuracy on adversarial samples perturbed from the original test dataset. Adversarial samples are crafted by $n$-step Projected Gradient Descent (PGD) with perturbation magnitude $\epsilon=8/255$ and step size $\alpha=2/255$ for both adversarial training and evaluation, with set $n=10$ for training and $n=20$ for evaluation, following~\citet{madry2017towards}.

\vspace{-1em}
\paragraph{Existing Baselines for Comparison.} We include several classical and competitive CIL approaches for comparison: \textbf{LwF-MC} is the multi-class implementation of~\citet{li2017learning}, and \textbf{LwF-MCMB} is its variation with a memory bank of previous data. For a fair comparison, it always keeps the same memory bank size with ours. For \textbf{iCaRL}~\citep{rebuffi2017icarl} and \textbf{IL2M}~\citep{Belouadah_2019_ICCV}, we use their publicly available implementations with the same memory bank sizes as ours. \textbf{IL2M} requires a second memory bank for storing previous class statistics. \textbf{DMC}~\citep{zhang2019class} is the existing CIL work that also utilizes external unlabeled data. We ensure DMC and CIL-QUD to query from the same external source.

\vspace{-1em}
\paragraph{Variants for Our Proposed Methods.} 
We first consider the vanilla version of CIL-QUD with neither queried unlabeled data nor the auxiliary classifier as Baseline among our proposals. It is still equipped with LwF regularizers yet only applied on stored labeled data\footnote{For a fair comparison, all settings are trained on the same amount of data in each training iteration. It means that the baseline approach can store extra historical data by taking over the ``budget'' for queried unlabeled data.} from previous learned tasks. We next check more self-ablations: 

\noindent \textit{\underline{a. CIL-QUD Variants:}} We study \textbf{four} variants of CIL-QUD explained as follows: {i)} Baseline + Auxiliary classifiers: it equals to CIL-QUD without involving the usage of unlabeled data; {ii)} CIL-QUD w. $\mathcal{FT}$; {iii)} CIL-QUD w. $\mathcal{KD}$; {iv)} CIL-QUD w. $\mathcal{KD}$ + CEM is the variant approach equipped with classifiers ensemble mechanism (CEM).

\noindent \textit{\underline{b. RCIL-QUD Variants:}} We investigate \textbf{ten} variants of RCIL-QUD with different robust LwF regularizations on queried unlabeled data, as listed in Table~\ref{table:adversarial_RA}. Classifier ensemble mechanism (CEM) is further applied to the top-performing settings. For all variants of CIL-QUD and RCIL-QUD without CEM, we evaluate the performance of the primary classifier by default, while the auxiliary classifier results are included in the supplement~\ref{sec:aux_res}. 

\vspace{-1em}
\paragraph{Privacy Issues of the Unlabeled Data Collection.} We believe that potential privacy concerns can be easily circumvented like by only querying from authorized and public datasets. To further inject privacy protection into our method, possible solutions include querying images with filtering of sensitive and offensive samples, or using generative replay: those are certainly feasible and can be our future works.

\vspace{-1em}
\paragraph{Ethical Issues of 80 Million Tiny Image Dataset.} We clarify that most of the experiments are completed in Spring 2020 before the withdrawal (June 2020) of 80 Million Tiny Image dataset.

Moreover, as demonstrated in Table~\ref{table:dif_source}, our approach can achieve a similar performance when it queries unlabeled data from 80 Million Tiny Image or ImageNet datasets. This provides an alternative and alleviates the potential ethical issues of adopting our methods for future researchers.

\vspace{-1mm}
\begin{elaboration}
\textbf{Takeaways:} Based on the results from Tables~\ref{table:standard} and~\ref{table:adversarial_RA}, we observe: ($1$) in terms of the standard accuracy, CIL-QUD w. $\mathcal{KD}$ + CEM establishes the SOTA performance; ($2$) in terms of the robust accuracy, CIL-QUD w. $\mathcal{RFT}$ + $\mathcal{RTC}$ + CEM reaches a superior performance.
\end{elaboration}
\vspace{-3mm}

\subsection{Improved Generalization with CIL-QUD}

\begin{wraptable}{r}{0.70\linewidth}
\begin{center}
\vspace{-1.7em}
\caption{\small Final performance for each task $\mathcal{T}$ of CIL-QUD, compared with SOTAs. Note that MT$_{\mathrm{upper}}$ and MT$_{\mathrm{lower}}$ train the model with multi-task learning scheme using full data and a few stored data, respectively. They act as the \textbf{empirical performance upper bounds and lower bounds} for CIL-QUD.}
\vspace{-4mm}
\label{table:standard}
\begin{threeparttable}
\resizebox{1\linewidth}{!}{
\begin{tabular}{l|c|c|c|c|c|c}
\toprule
\multirow{2}{*}{Methods} & \multicolumn{6}{c}{CIFAR-10 (SA)} \\ \cmidrule{2-7} 
 & $\mathcal{T}_1$ (\%) & $\mathcal{T}_2$ (\%) & $\mathcal{T}_3$ (\%) & $\mathcal{T}_4$ (\%) & $\mathcal{T}_5$ (\%) & Average (\%)\\ \midrule
LwF-MC~\citep{li2017learning} & 28.30 & 58.10 & 50.20 & 46.00 & 60.25 & 48.57\\
LwF-MCMB~\citep{li2017learning} & 76.45 & 81.45 & 69.55 & 32.30 & 44.30 & 60.81\\
iCaRL~\citep{rebuffi2017icarl} & 76.45 & 79.00 & 75.70 & 50.85 & 35.55 & 63.51\\
IL2M~\citep{Belouadah_2019_ICCV} & 78.20 & 64.05 & 60.40 & 38.95 & 92.10 & 66.74\\
DMC~\citep{zhang2019class} & - & - & - & - & - & \RV{73.66} \\
\midrule
MT$_{\mathrm{lower}}$ (empirical lower bound) & 62.75 & 59.30 & 51.10 & 37.40 & 49.95 & 52.10\\
MT$_{\mathrm{upper}}$  (empirical uppper bound) & 95.75 & 95.75 & 91.60 & 89.10 & 92.30 & 92.90\\
\midrule
Vanilla Baseline & 78.05 & 64.05 & 59.75 & 36.75 & 92.40 & 66.20\\
Baseline + Auxiliary Classifier & 75.05 & 71.50 & 54.25 & 52.05 & 89.10 & 68.39\\
CIL-QUD w. $\mathcal{FT}$ & 83.10 & 80.40 & 63.50 & 75.40 & 62.15 & 72.91 \\
CIL-QUD w. $\mathcal{KD}$ & 82.05 & 77.55 & 72.05 & 66.15 & 74.05 & 74.37\\
CIL-QUD w. $\mathcal{KD}$ + CEM & 84.40 & 79.50 & 69.40 & 66.85 & 84.95 & 77.02 \\
\toprule
\multirow{2}{*}{Methods} & \multicolumn{6}{c}{CIFAR-100 (SA)} \\ \cmidrule{2-7}
 & $\mathcal{T}_1$ (\%) & $\mathcal{T}_2$ (\%) & $\mathcal{T}_3$ (\%) & $\mathcal{T}_4$ (\%) & $\mathcal{T}_5$ (\%) & Average (\%)\\ \midrule
IL2M~\citep{Belouadah_2019_ICCV} & 22.05 & 18.70 & 33.45 & 32.35 & 82.90 & 37.89\\
DMC~\citep{zhang2019class} & 40.32 & 39.75 & 42.10 & 46.12 & 59.36 & 45.53\\
\midrule
MT$_{\mathrm{lower}}$  (empirical lower bound) & 14.40 & 10.50 & 12.30 & 11.15 & 16.75 & 13.02\\
MT$_{\mathrm{upper}}$  (empirical upper bound) & 70.65 & 67.55 & 75.05 & 68.80 & 74.55 & 71.32 \\
\midrule
CIL-QUD w. $\mathcal{KD}$ & 30.85 & 39.30 & 52.65 & 41.80 & 67.20 & 46.36 \\
CIL-QUD w. $\mathcal{KD}$ + CEM & 51.95 & 46.15 & 52.80 & 38.60 & 44.10 & 46.72\\
\bottomrule
\end{tabular}}
\end{threeparttable}
\end{center}
\vspace{-5mm}
\end{wraptable}

\begin{figure}[!ht]
    \vspace{-2mm}
    \centering
    \includegraphics[width=1\linewidth]{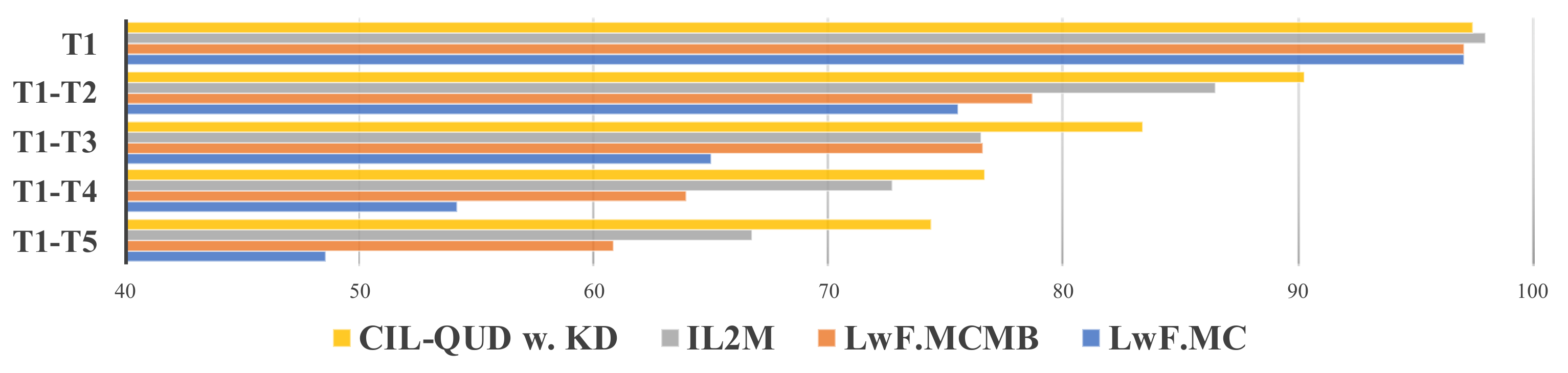}
    \vspace{-2em}
    \caption{\small Average performance variation when training CIL models incrementally over $5$ tasks on the CIFAR-10 dataset with selected approaches. T$_1$-T$_i$ means the model have been trained with tasks $1,\cdots,i$ ($i\in\{1,2,3,4,5\}$) continually.}
    \label{fig:res}
    \vspace{-4mm}
\end{figure}

The standard accuracies (SAs) of all CIL models are collected in this section. As shown in Table~\ref{table:standard}, several observations could be drawn: (1) By incrementally adding our proposed components to the vanilla baseline, we find all of them contribute to preserving SA. Among the auxiliary classifier architecture (+$2.37\%$ SA), unlabeled data regularization $\mathcal{L}_{\mathrm{LwF}}$ (+$4.52\%$ SA from $\mathcal{FT}$; +$5.98\%$ SA from $\mathcal{KD})$, and CEM (+$2.65\%$), $\mathcal{L}_{\mathrm{LwF}}$ is the dominant contributor, which demonstrates that queried external unlabeled data gently balances the asymmetric information between previous classes and newly added classes in CIL. (2) Comparing CIL-QUD w. $\mathcal{KD}$ with CIL-QUD w. $\mathcal{FT}$, $\mathcal{KD}$ seems to yield obtains a larger SA boost than $\mathcal{FT}$. (3) All results in Table~\ref{table:standard} show CIL-QUD significantly outperforms previous SOTA methods by substantial margins on both CIFAR-10 (+$10.28\%$ SA) and CIFAR-100 (at least +$1.19\%$ SA). It demonstrates that, aided by query unlabeled data, CIL-QUD heavily reduces the prediction bias towards either the previous or the new classes. Note that our proposal also surpasses DMC~\citep{zhang2019class} by $1.19\%$ SA on CIFAR-100, with the same amount of unlabeled data: that supplies evidence that anchor-based query is more effective than blindly sampling. In addition, we report the average performance variation when incrementally training CIL models in Figure~\ref{fig:res}. We observe that CIL-QUD w. $\mathcal{KD}$ stably does better in preserving previous knowledge besides superior final performance. 

\subsection{Improved Robustness with RCIL-QUD} \label{sec:rcil}

\begin{wraptable}{r}{0.70\linewidth}
\begin{center}
\vspace{-0.5em}
\caption{\small Robust accuracy for each task $\mathcal{T}_i$ of RCIL-QUD and its variants. MTAT$_{\mathrm{upper}}$ and MTAT$_{\mathrm{lower}}$ adversarially train the model with multi-task learning scheme using full data and a few stored data, respectively. They act as the \textbf{empirical performance upper bounds and lower bounds} for RCIL-QUD. Here we focus on reporting RAs, while their corresponding SAs are in Table~\ref{table:adversarial_SA}.}
\vspace{-4mm}
\label{table:adversarial_RA}
\begin{threeparttable}
\resizebox{1\linewidth}{!}{
\begin{tabular}{l|c|c|c|c|c|c}
\toprule
\multirow{2}{*}{Methods} & \multicolumn{6}{c}{CIFAR-10 (RA)} \\ \cmidrule{2-7}
 & \multicolumn{1}{c|}{$\mathcal{T}_1$ (\%)} & \multicolumn{1}{c|}{$\mathcal{T}_2$ (\%)} & \multicolumn{1}{c|}{$\mathcal{T}_3$ (\%)} & \multicolumn{1}{c|}{$\mathcal{T}_4$ (\%)} & \multicolumn{1}{c|}{$\mathcal{T}_5$ (\%)} & \multicolumn{1}{c}{Average (\%)}\\ \midrule
MTAT$_{\mathrm{lower}}$ & 22.18 & 15.99 & 18.42 & 4.29 & 7.66 & 13.71\\
MTAT$_{\mathrm{upper}}$ & 54.00 & 58.70 & 39.60 & 28.30 & 21.35 & 40.39\\
\midrule
RCIL-QUD w. $\mathcal{KD}$ & 4.10 & 14.30 & 6.35 & 5.10 & 24.20 & 10.81\\
RCIL-QUD w. $\mathcal{RKD}$ & 21.00 & 37.90 & 29.25 & 3.90 & 16.30 & 21.67\\
RCIL-QUD w. $\mathcal{RKD}$ + CEM & 34.70 & 45.10 & 25.50 & 11.60 & 36.95 & 30.77 \\
RCIL-QUD w. $\mathcal{RKD}$ + $\mathcal{RTC}$ & 37.00 & 30.00 & 30.60 & 16.60 & 5.10 & 23.86\\
RCIL-QUD w. $\mathcal{RKD}$ + $\mathcal{RTC}$ + CEM & 48.15 & 50.85 & 39.40 & 19.60 & 42.75 & \textcolor{red}{40.15}\\
\midrule
RCIL-QUD w. $\mathcal{FT}$ & 21.70 & 26.65 & 11.90 & 6.45 & 8.85 & 15.11\\
RCIL-QUD w. $\mathcal{RFT}$ & 25.45 & 34.05 & 19.90 & 5.15 & 6.05 & 18.12 \\
RCIL-QUD w. $\mathcal{RFT}$ + CEM & 26.75 & 35.10 & 21.50 & 6.85 & 23.80 & 22.80\\
RCIL-QUD w. $\mathcal{RFT}$ + $\mathcal{RTC}$ & 32.45 & 29.55 & 28.10 & 23.12 & 10.13 & 24.67\\
RCIL-QUD w. $\mathcal{RFT}$ + $\mathcal{RTC}$ + CEM & 41.45 & 37.20 & 31.95 & 16.35 & 24.30 & 30.25\\
\bottomrule
\toprule
\multirow{2}{*}{Methods} & \multicolumn{6}{c}{CIFAR-100 (RA)} \\ \cmidrule{2-7}
 & \multicolumn{1}{c|}{$\mathcal{T}_1$ (\%)} & \multicolumn{1}{c|}{$\mathcal{T}_2$ (\%)} & \multicolumn{1}{c|}{$\mathcal{T}_3$ (\%)} & \multicolumn{1}{c|}{$\mathcal{T}_4$ (\%)} & \multicolumn{1}{c|}{$\mathcal{T}_5$ (\%)} & \multicolumn{1}{c}{Average (\%)}\\ \midrule
MTAT$_{\mathrm{lower}}$ & 3.65 & 2.25 & 1.85 & 1.75 & 5.30 & 2.96\\
MTAT$_{\mathrm{upper}}$ & 23.40 & 16.35 & 20.40 & 19.20 & 29.05 & 21.68 \\
\midrule
RCIL-QUD w. $\mathcal{RKD}$ & 9.50 & 5.05 & 10.10 & 13.00 & 24.50 & 12.43 \\
RCIL-QUD w. $\mathcal{RKD}$ +CEM & 15.10 & 8.60 & 13.20 & 15.55 & 21.45 & 14.78\\
RCIL-QUD w. $\mathcal{RFT}$ & 5.00 & 3.45 & 6.30 & 7.65 & 26.90 & 9.86 \\
RCIL-QUD w. $\mathcal{RFT}$ + CEM & 12.05 & 7.05 & 10.95 & 10.15 & 19.70 & 11.98\\
\bottomrule 
\end{tabular}}
\end{threeparttable}
\end{center}
\vspace{-6mm}
\end{wraptable}

Under the new robustness-aware CIL scheme, the robust accuracies (RAs) of all models are collected in this section. As shown in Table~\ref{table:adversarial_RA} and~\ref{table:adversarial_SA}, a number of observations can be drawn consistently on CIFAR-10 and CIFAR-100 datasets. 

\underline{First}, $\mathcal{FT}$ regularizer surpasses $\mathcal{KD}$ by $12.13\%$ SA and $4.3\%$ RA on CIFAR-10, which supports~\citet{shafahi2019adversarially,boopathy2020proper}'s claim that regularizing feature presentations are crucial for robustness. Compared to our empirical performance lower bound by MTAT$_{\mathrm{lower}}$ (explained in Table~\ref{table:adversarial_RA} caption), $\mathcal{KD}$ even hurts RA. A possible explanation is that ill-conditioned probabilities may obstacle preserving robustness. \underline{Second}, robustified regularizers on unlabeled data, $\mathcal{RKD}$ and $\mathcal{RFT}$, significantly increase RA compare with their standard versions, especially for $\mathcal{RKD}$ (+$11.03\%$ SA and +$10.86\%$ RA on CIFAR-10). It suggests that injecting adversarial signals into regularizations effectively improves CIL models' resilience. \underline{Third}, $\mathcal{RTC}$ further pushes RA higher, with slight TA degradation. As the \textcolor{red}{red} RA number shows in Table~\ref{table:adversarial_RA}, it only has a negligible gap ($0.24\%$) to the empirical RA upper bound from MTAT$_{\mathrm{upper}}$, which we consider as an almost insurmountable milestone for achievable robustness in CIL. \underline{Lastly}, the classifier ensemble mechanism (CEM) continues to benefit RAs on both CIFAR-10 and CIFAR-100. 

\subsection{Ablation Study}

\begin{wrapfigure}{r}{0.50\linewidth}
    \centering
    \vspace{-4mm}
    \includegraphics[width=0.95\linewidth]{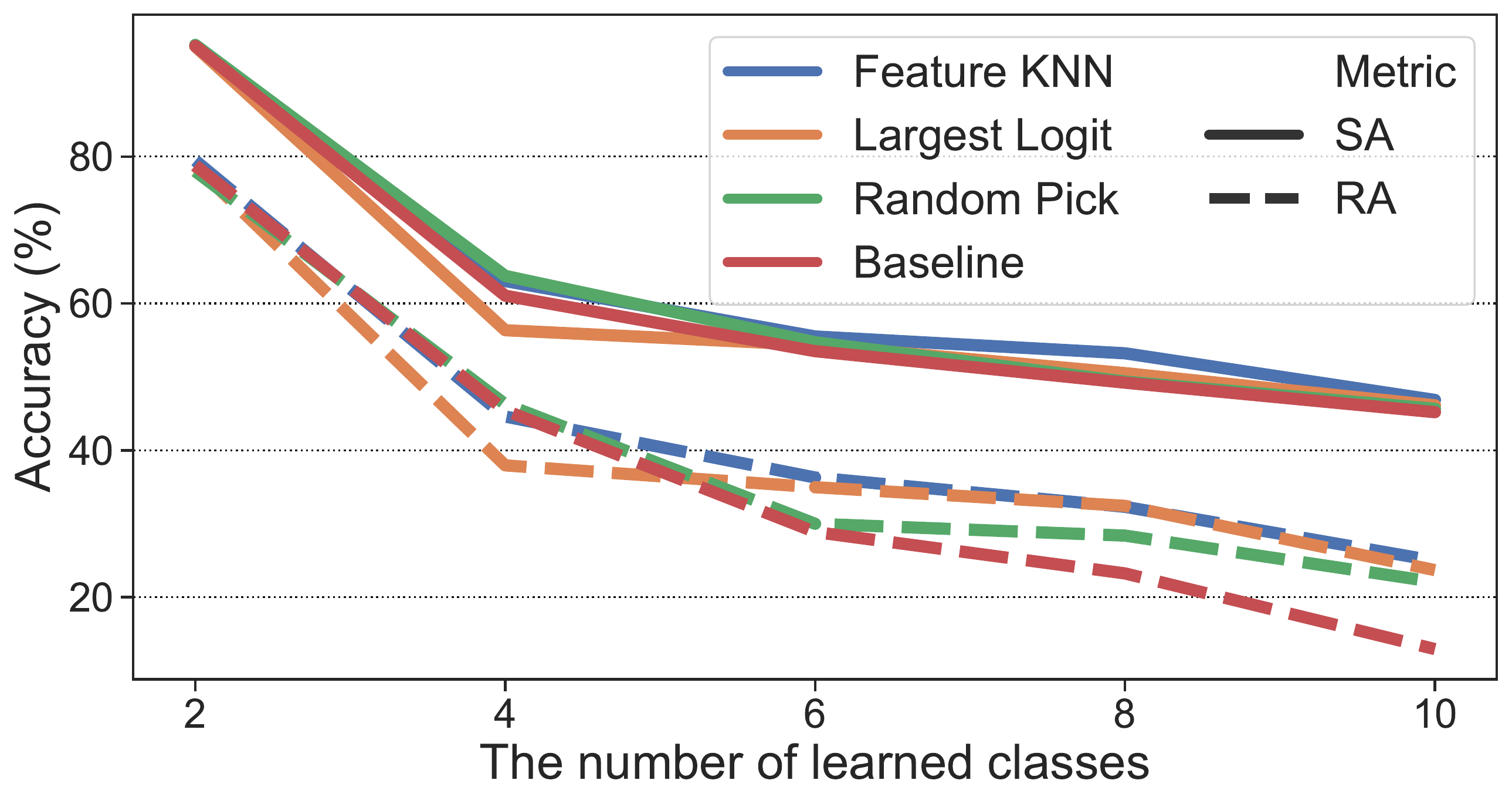}
    \vspace{-4mm}
    \caption{\small Average performance of RCIL-QUD w. $\mathcal{RFT}$ + $\mathcal{RTC}$ on CIFAR-10 with different unlabeled data query methods. Both RAs and corresponding SAs are reported.}
    \label{fig:query}
    \vspace{-8mm}
\end{wrapfigure}

\vspace{-1mm}
\paragraph{Do we indeed need query, and would random unlabeled data also work?} CIL-QUD has already outperformed DMC~\citep{zhang2019class} in Section 4.2, which endorses the query way for better SAs. We further conduct experiments to justify our query also benefits RAs in RCIL-QUD w. $\mathcal{RFT}$ + $\mathcal{RTC}$. We first observe that leveraging queried unlabeled data leads to better generalization and robustness than just storing a few historical data from past tasks (i.e., Baseline in Figure~\ref{fig:query}). Then, we systematically compare the effectiveness of the following methods on utilizing unlabeled data: (i) Feature KNN: finding K nearest-neighbors for each stored anchor over feature embeddings; which is our default way; (ii) Largest Logit: finding unlabeled data with top K largest soft logits for each class; (iii) Random Pick: randomly sampling unlabeled data. 

As shown in Figure~\ref{fig:query}, all three options utilizing unlabeled data steadily outperform the vanilla baseline, especially in terms of RA. Our default query mechanism with feature KNN performs the best.

\paragraph{The influence of ``anchors'' for query.} We adopt two representative activate learning methods to choose the ``anchors'': (1) maximum entropy sampling~\citep{lewis1994sequential,settles2009active} termed as ``Entropy''; and (2) core-set selection via proxy~\citep{coleman2019selection} named as ``SVP''. Experiments of CIL-QUD w. $\mathcal{KD}$ on CIFAR-10 have been carried out, with the settings from Table~\ref{table:standard}. We obtain: Random v.s. Entropy v.s. SVP = $74.37\%$ v.s. $73.91\%$ v.s. $74.76\%$. Results imply that different choices of ``anchors'' have limited effects on the achievable performance of our methods.

\begin{table}[!ht]
\begin{minipage}{0.48\linewidth}
\caption{\small Investigation of diverse sizes of the memory bank $\mathcal{S}$.}
\vspace{-5mm}
\label{table:memorybank}
\begin{center}
\resizebox{0.95\textwidth}{!}{
\begin{tabular}{c|c|c|c|c|c|c}
\toprule
\multirow{2}{*}{Size per class} & \multicolumn{6}{c}{CIL-QUD (Standard Accuracy) on CIFAR-10} \\ \cmidrule{2-7} 
 & $\mathcal{T}_1$ (\%) & $\mathcal{T}_2$ (\%) & $\mathcal{T}_3$ (\%) & $\mathcal{T}_4$ (\%) & $\mathcal{T}_5$ (\%) & Average (\%)\\ \midrule
10 & 65.15 & 63.20 & 57.55 & 56.70 & 67.70 & 62.06\\
20 & 74.90 & 72.45 & 61.30 & 71.20 & 61.40 & 68.25\\
50 & 82.30 & 71.75 & 66.35 & 63.65 & 71.10 & 71.03\\
100 & 82.05 & 77.55 & 72.05 & 66.15 & 74.05 & 74.37\\
\bottomrule
\toprule
\multirow{2}{*}{Size per calss} & \multicolumn{6}{c}{RCIL-QUD (Robust Accuracy) on CIFAR-10} \\ \cmidrule{2-7} 
 & $\mathcal{T}_1$ (\%) & $\mathcal{T}_2$ (\%) & $\mathcal{T}_3$ (\%) & $\mathcal{T}_4$ (\%) & $\mathcal{T}_5$ (\%) & Average (\%)\\ \midrule
10 & 24.16 & 27.64 & 16.98 & 11.40 & 6.32 & 17.30\\
20 & 28.95 & 27.90 & 23.30 & 20.65 & 8.65 & 21.89\\
50 & 31.35 & 28.15 & 26.20 & 21.05 & 9.44 & 23.24\\
100 & 32.45 & 29.55 & 28.10 & 23.12 & 10.13 & 24.67\\
\bottomrule
\end{tabular}}
\end{center}
\end{minipage}
\begin{minipage}{0.48\linewidth}  
\caption{\small Query unlabeled data from different sources.}
\vspace{-5mm}
\label{table:dif_source}
\begin{center}
\resizebox{0.95\textwidth}{!}{
\begin{tabular}{c|c|c|c|c|c|c}
\toprule
\multirow{2}{*}{Sources} & \multicolumn{6}{c}{CIL-QUD (Standard Accuracy) on CIFAR-10} \\ \cmidrule{2-7} 
 & $\mathcal{T}_1$ (\%) & $\mathcal{T}_2$ (\%) & $\mathcal{T}_3$ (\%) & $\mathcal{T}_4$ (\%) & $\mathcal{T}_5$ (\%) & Average (\%)\\ \midrule
80 Million & 82.05 & 77.55 & 72.05 & 66.15 & 74.05 & 74.37\\
ImageNet & 78.25 & 72.45 & 68.20 & 73.80 & 70.15 & 72.57\\
\bottomrule
\toprule
\multirow{2}{*}{Sources} & \multicolumn{6}{c}{RCIL-QUD (Robust Accuracy) on CIFAR-10} \\ \cmidrule{2-7} 
 & $\mathcal{T}_1$ (\%) & $\mathcal{T}_2$ (\%) & $\mathcal{T}_3$ (\%) & $\mathcal{T}_4$ (\%) & $\mathcal{T}_5$ (\%) & Average (\%)\\ \midrule
80 Million & 32.45 & 29.55 & 28.10 & 23.12 & 10.13 & 24.67 \\
ImageNet & 34.35 & 27.90 & 26.03 & 18.75 & 9.37 & 23.28\\
\bottomrule
\end{tabular}}
\end{center}
\end{minipage}
\end{table}

\paragraph{Changing the size of memory bank.} Indicated by~\citet{rebuffi2017icarl,Belouadah_2019_ICCV}, the size of memory back $S$ plays a key role in the CIL performance. We conduct an ablation of CIL-QUD w. $\mathcal{KD}$ and RCIL-QUD w. $\mathcal{RFT}$ + $\mathcal{RTC}$ with $10, 20, 50, 100$ samples stored per class, as shown in Table~\ref{table:memorybank}. Note that we keep the number of queried unlabeled data the same default, across all settings for a fair comparison. We observe that CIL-QUD/RCIL-QUD equipped with a large memory banks consistency achieve the superior performance, in term of both generalization and robustness.  That is understandable that more anchors ensure more relevant queried data. Meanwhile, we are also encouraged to see CIL-QUD and RCIL-QUD already have competitive performance when each class has as small as 20-50 stored samples, demonstrating a nice sample and memory efficiency of our methods.

\paragraph{Switching to different sources of unlabeled data} Since the 80 Million Tiny Image set is often considered to be from the same distribution with CIFAR-10/-100, we switch the unlabeled data source to ImageNet ($32\times32$ downsampled versions), whose images are semantically more complex and noisy. Results are presented in Table~\ref{table:dif_source}. We observe that the achieved performance of CIL-QUD/RCIL-QUD stay robust under a different unlabeled data distribution.

\section{Conclusions}
\vspace{-0.3em}
We introduce unlabeled data queries and show that it effectively overcome the catastrophic forgetting in the standard CIL scheme, with the help of LwF regularizations. Through anchor-based query, unlabeled data brings in more relevant information to previous classes while alleviating the information asymmetry. The power of queried unlabeled data can further extend to preserving robustness in CIL, opening up a new dimension of robustified CIL and establishing strong milestone results. Our results exemplify a significant reduction of the performance gap between incremental and non-incremental learning, in terms of both SA and RA. Similarly to other deep learning frontiers, unlabeled data evidently reveals the new promise for CIL too, if leveraged appropriately. We hope that our findings can broadly inspire people to dig deeper into utilizing unlabeled data for lifelong learning, knowledge transfer, and sustained robustness. 

\section*{Acknowledgement}
\vspace{-0.3em}

Z.W. is in part supported by the U.S. Army Research Laboratory Cooperative Research Agreement W911NF17-2-0196 (IOBT REIGN), and an IBM faculty research award.


\bibliography{AdvCL}
\bibliographystyle{tmlr}

\clearpage

\appendix
\renewcommand{\thepage}{S\arabic{page}}  
\renewcommand{\thesection}{S\arabic{section}}   
\renewcommand{\thetable}{S\arabic{table}}   
\renewcommand{\thefigure}{S\arabic{figure}}

\section{More Methodology Details}

\subsection{Classifier Ensemble Mechanism Details} \label{sec:SEM}
Here we present the details of our \textbf{C}lassifier \textbf{E}nsemble \textbf{M}echanism (CEM) mentioned in Section~\ref{balance_training}. We follow the same notations as those provided in the main text. Note that both CIL-QUD and RCIL-QUD produce a primary classifier $\mathcal{M}^{\mathrm{PC}}(\boldsymbol{\theta}_{\mathrm{c},1},\varphi(\boldsymbol{\theta},\mathbf x))$ and an auxiliary classifier $\mathcal{M}^{\mathrm{AC}}(\boldsymbol{\theta}_{\mathrm{c},2},\varphi(\boldsymbol{\theta},\mathbf x))$ when the current incremental learning is finished. Although both classifiers give satisfying results, there exists still a performance trade-off~\citep{zhang2019balance} between them, which motivates us to propose CEM for extra performance improvement. Specifically, for an inference image, we select its K-nearest-neighbors (in our case, K$=50$) from current accessible stored data and queried unlabeled data via their feature embedding. Then, a majority voting is applied with selected nearest neighbors to obtain a ``predicted task ID'' for the inference image. If the image belongs to previous tasks, we employ $\mathcal{M}^{\mathrm{PC}}$ to classify; otherwise, $\mathcal{M}^{\mathrm{AC}}$ is chosen for predictions. The critical point here is the subtle usage of hidden information ($e.g.$ task ID) within stored data and queried unlabeled data, which assists the inference. The algorithm~\ref{alg:vote} outlines the full procedure of CEM.

\begin{algorithm}
\SetAlgoLined
\KwIn{An inference image $x$, the memory bank $\mathcal{S}=\{\mathcal{S}_1,\cdots,\mathcal{S}_n\}$ including $n$ previous tasks, the queried unlabeled data $\mathcal{U}=\{\mathcal{U}_1,\cdots,\mathcal{U}_n\}$, the trained CIL-QUD or RCIL-QUD feature extractor $\varphi$ after learning $n$ tasks continually, a primary classifier $\mathcal{M}^{\mathrm{PC}}=(\mathcal{M}^{\mathrm{PC}}_1,\cdots,\mathcal{M}^{\mathrm{PC}}_n)$, and an auxiliary classifier $\mathcal{M}^{\mathrm{AC}}=(\mathcal{M}^{\mathrm{AC}}_1,\cdots,\mathcal{M}^{\mathrm{AC}}_n)$}
\KwOut{A prediction label $\hat{y}$}
Distance set $D \leftarrow \emptyset$\\
Compute the feature embedding $\boldsymbol{e}^*=\varphi(\mathbf x)$ for inference image $\mathbf x$\\
\For{$i \gets 1$ \KwTo $n$}
{
    \For{each image $\tilde{\mathbf x}$ in $\mathcal{S}_i\bigcup\mathcal{U}_i$}
    {
        Compute Euclidean Distance $d$ between $\varphi(\tilde{\mathbf x})$ and $e^*$\\
        $D=D\bigcup\{(d,i)\}$, where $i$ is the task ID\\
    }
}
Sort and select the top-K tuples in $D$ \\
Picking the majority task ID $j$ of top-K tuples\\
\uIf{j equal to $n$}{
$\hat{y}=\argmax\mathcal{M}^{\mathrm{AC}}_n(\boldsymbol{e}^*)$
}
\Else{
$\hat{y}=\argmax\mathcal{M}^{\mathrm{PC}}_j(\boldsymbol{e}^*)$
}
\caption{Classifier Ensemble Mechanism}
\label{alg:vote}
\end{algorithm}

\section{More Implementation Details} \label{sec:setup}
\paragraph{Sources for Queried Unlabeled Data.} All queried unlabeled data are from of 80 Million Tiny Image dataset~\citep{torralba200880} or ImageNet dataset ($32\times 32$ downsampled versions)~\citep{deng2009imagenet}.

\paragraph{Training Details and Model Picking.} In our case, all models are trained using Stochastic Gradient Descent (SGD) with $0.9$ momentum. On the CIFAR-10 dataset, for $100$ epochs standard training, our learning rate starts from $0.01$ and decays by $10$ times at epochs $60$ and $80$; for $80$ epochs adversarial training, our learning rate starts from $0.01$ and decays by $10$ times at epochs $40$ and $60$. On CIFAR-100 dataset, both standard and adversarial training have $40$ epochs, where the learning rate starts from $0.01$ and decays by $10$ times at epochs $20$ and $30$. We choose a total batch size to be $256$ in our case. Specifically, we adopt $64$ as a class-balanced batch size, $64$ as randomly sampling batch size, and $128$ as unlabeled data batch size. In this paper, we pick trained models with the highest standard accuracy on the holdout validation dataset.

\paragraph{More Methods for Comparisons.} Moreover, the necessary baselines of performance lower bound and upper bound are also considered. MT$_{\mathrm{lower}}$ trains the model only using stored data ($1,000$ images in our case) for all classes with the multi-task learning scheme, which is the lower bound of performance for CIL-QUD. MT$_{\mathrm{upper}}$ trains the model using full data for all classes with the multi-task learning scheme, which is the upper bound of performance for CIL-QUD. MTAT$_{\mathrm{lower}}$ performs AT only using stored labeled data ($1,000$ images in our case) for all classes with the multi-task learning scheme, which is the lower bound of performance for RCIL-QUD. MTAT$_{\mathrm{upper}}$ performs AT using full labeled data with the multi-task learning scheme and combining TRADES on all queried unlabeled data\footnote{For a fair comparison, we query the same number of unlabeled data for all experiments.}, which is the performance upper bound for RCIL-QUD.

\section{More Experiment Results}

\subsection{Combine CIL-QUD with SOTA method IL2M}

\begin{table}[ht]
\begin{center}
\caption{Final performance for each task $\mathcal{T}_i$ from CIL-QUD + IL2M. PC: Primary Classifier; AC: Auxiliary Classifier.}
\vspace{-3mm}
\label{table:standard_2}
\begin{threeparttable}
\resizebox{0.8\textwidth}{!}{
\begin{tabular}{l|c|c|c|c|c|c}
\toprule
\multirow{2}{*}{Methods} & \multicolumn{6}{c}{CIFAR-10 (SA)} \\ \cmidrule{2-7} 
 & $\mathcal{T}_1$ (\%) & $\mathcal{T}_2$ (\%) & $\mathcal{T}_3$ (\%) & $\mathcal{T}_4$ (\%) & $\mathcal{T}_5$ (\%) & Average (\%)\\ \midrule
IL2M & 78.20 & 64.05 & 60.40 & 38.95 & 92.10 & 66.74\\ \midrule
CIL-QUD w. $\mathcal{KD}$ (PC) & 82.05 & 77.55 & 72.05 & 66.15 & 74.05 & 74.37\\
CIL-QUD w. $\mathcal{KD}$ + IL2M (PC) & 82.05 & 77.55 & 72.05 & 66.15 & 74.05 & 74.37\\ \midrule
CIL-QUD w. $\mathcal{KD}$ (AC) & 44.00 & 51.50 & 22.10 & 26.40 & 95.05 & 47.81 \\
CIL-QUD w. $\mathcal{KD}$ + IL2M (AC) & 51.20 & 58.90 & 41.40 & 68.65 & 91.20 & 62.27 \\
\bottomrule
\end{tabular}}
\end{threeparttable}
\end{center}
\vspace{-1em}
\end{table}

Our framework with regularizers on queried unlabeled data can be formulated as a Plug-and-Play mechanism and flexibly combined with existing CIL approaches. As shown in Table~\ref{table:standard_2}, we present the results of the variant approach, CIL-QUD w. $\mathcal{KD}$ + IL2M, which combines our framework CIL-QUD w. $\mathcal{KD}$ with the previous state-of-the-art (SOTA) method IL2M~\citep{Belouadah_2019_ICCV}. The performances of both primary classifiers and auxiliary classifiers have been collected. From Table~\ref{table:standard_2}, IL2M contributes an extra performance boost based on CIL-QUD, while it mainly strengthens the power of auxiliary classifiers. It should come as no surprise since IL2M focuses on compensating the prediction bias introduced by imbalanced data training, which exactly resolves the limitation of auxiliary classifiers. This is also the reason why IL2M appears to be useless for the primary classifier, which is trained with class-balanced data batches.

\subsection{Auxiliary Classifier Results} \label{sec:aux_res}
In this section, we report the auxiliary classifiers' performance of several important CIL-QUD and RCIL-QUD variants. As shown in Table~\ref{table:std_aux} and Table~\ref{table:adv_aux}, the results demonstrate that the auxiliary classifier, trained with randomly sampled data batches, achieves high SA and RA on the majority classes (newly added classes) while tends to forget the knowledge of minority classes (previous classes). As for average performance, most of the auxiliary classifiers perform worse than primary classifiers, which have shown strong abilities to inherit previous classes' information.

\begin{table}[!ht]
\begin{center}
\caption{Final performance for each task $\mathcal{T}_i$ from Auxiliary Classifiers of CIL-QUD.}
\vspace{-3mm}
\label{table:std_aux}
\begin{threeparttable}
\resizebox{0.8\textwidth}{!}{
\begin{tabular}{l|c|c|c|c|c|c}
\toprule
\multirow{2}{*}{Methods} & \multicolumn{6}{c}{CIFAR-10 (SA)} \\ \cmidrule{2-7} 
 & $\mathcal{T}_1$ (\%) & $\mathcal{T}_2$ (\%) & $\mathcal{T}_3$ (\%) & $\mathcal{T}_4$ (\%) & $\mathcal{T}_5$ (\%) & Average (\%)\\ \midrule
CIL-QUD w. $\mathcal{FT}$ & 62.20 & 50.10 & 26.90 & 21.05 & 95.95 & 51.24 \\
CIL-QUD w. $\mathcal{KD}$ & 43.35 & 45.50 & 29.45 & 34.15 & 94.75 & 49.44 \\
\bottomrule
\end{tabular}}
\end{threeparttable}
\end{center}
\end{table}

\begin{table}[!ht]
\begin{center}
\caption{Final performance for each task $\mathcal{T}_i$ from Auxiliary Classifiers of RCIL-QUD.}
\vspace{-3mm}
\label{table:adv_aux}
\begin{threeparttable}
\resizebox{0.8\textwidth}{!}{
\begin{tabular}{l|c|c|c|c|c|c}
\toprule
\multirow{2}{*}{Methods} & \multicolumn{6}{c}{CIFAR-10 (SA)} \\ \cmidrule{2-7} 
 & $\mathcal{T}_1$ (\%) & $\mathcal{T}_2$ (\%) & $\mathcal{T}_3$ (\%) & $\mathcal{T}_4$ (\%) & $\mathcal{T}_5$ (\%) & Average (\%)\\ \midrule
RCIL-QUD w. $\mathcal{RKD}$ & 19.80 & 35.45  & 20.45  & 6.40  & 91.95 & 34.81  \\
RCIL-QUD w. $\mathcal{RKD}$ + $\mathcal{RTC}$ & 4.15 & 26.30 & 6.10  & 0.55  & 88.30  & 25.08 \\
\hline
RCIL-QUD w. $\mathcal{RFT}$ & 25.40 & 40.20  & 9.60 & 4.30  & 93.45 & 34.59 \\
RCIL-QUD w. $\mathcal{RFT}$ + $\mathcal{RTC}$ & 2.60 & 2.85 & 0.20 & 0.00 & 79.20 & 16.97 \\
\bottomrule
\toprule
\multirow{2}{*}{Methods} & \multicolumn{6}{c}{CIFAR-10 (RA)} \\ \cmidrule{2-7} 
 & $\mathcal{T}_1$ (\%) & $\mathcal{T}_2$ (\%) & $\mathcal{T}_3$ (\%) & $\mathcal{T}_4$ (\%) & $\mathcal{T}_5$ (\%) & Average (\%)\\ \midrule
RCIL-QUD w. $\mathcal{RKD}$ & 3.85 & 14.00 & 7.70 & 0.60 & 58.05 & 16.84 \\
RCIL-QUD w. $\mathcal{RKD}$ + $\mathcal{RTC}$ & 0.60 & 13.75 & 1.90 & 0.00 & 65.50 & 16.35 \\
\hline
RCIL-QUD w. $\mathcal{RFT}$ & 3.35 & 11.10 & 1.00 & 0.10  & 48.50 & 12.81 \\
RCIL-QUD w. $\mathcal{RFT}$ + $\mathcal{RTC}$ & 0.40 & 1.10 & 0.00 & 0.00 & 62.70 & 12.84 \\
\bottomrule
\end{tabular}}
\end{threeparttable}
\end{center}
\end{table}

\subsection{More Results of RCIL-QUD}

Table~\ref{table:adversarial_SA} reports the standard accuracy for each task $\mathcal{T}_i$ of RCIL-QUD and its variants, which provides results for the analyses in Section~\ref{sec:rcil}. 

\begin{table}[!ht]
\begin{center}
\caption{\small Standard accuracy for each task $\mathcal{T}_i$ of RCIL-QUD and its variants. MTAT$_{\mathrm{upper}}$ and MTAT$_{\mathrm{lower}}$ adversarially trains the model with multi-task learning scheme using full data and a few stored data, respectively. They offers the empirical performance upper bound and lower bound for RCIL-QUD.}
\vspace{-3mm}
\label{table:adversarial_SA}
\begin{threeparttable}
\resizebox{0.8\textwidth}{!}{
\begin{tabular}{l|c|c|c|c|c|c}
\toprule
\multirow{2}{*}{Methods} & \multicolumn{6}{c}{CIFAR-10 (SA)} \\ \cline{2-7} 
 & \multicolumn{1}{c|}{$\mathcal{T}_1$ (\%)} & \multicolumn{1}{c|}{$\mathcal{T}_2$ (\%)} & \multicolumn{1}{c|}{$\mathcal{T}_3$ (\%)} & \multicolumn{1}{c|}{$\mathcal{T}_4$ (\%)} & \multicolumn{1}{c|}{$\mathcal{T}_5$ (\%)} & \multicolumn{1}{c}{Average (\%)}\\ \midrule
MTAT$_{\mathrm{lower}}$ & 57.18 & 50.97 & 45.93 & 31.02 & 40.83 & 45.19 \\
MTAT$_{\mathrm{upper}}$ & 91.50 & 89.90 & 79.30 & 83.35 & 72.00 & 83.21 \\
\midrule
RCIL-QUD w. $\mathcal{KD}$ & 38.70 & 48.75 & 34.35 & 48.70 & 83.60 & 50.82\\
RCIL-QUD w. $\mathcal{RKD}$ & 61.35 & 75.50 & 70.85 & 35.20 & 66.35 & 61.85\\
RCIL-QUD w. $\mathcal{RKD}$ + CEM & 69.95 & 74.40 & 52.90 & 44.10 & 78.90 & 64.05\\
RCIL-QUD w. $\mathcal{RKD}$ + $\mathcal{RTC}$ & 67.00 & 60.50 & 56.95 & 43.40 & 18.40 & 49.25\\
RCIL-QUD w. $\mathcal{RKD}$ + $\mathcal{RTC}$ + CEM & 70.00 & 67.70 & 53.80 & 37.35 & 66.85 & 59.14\\
\midrule 
RCIL-QUD w. $\mathcal{FT}$ & 68.60 & 69.65 & 48.20 & 63.85 & 64.45 & 62.95\\
RCIL-QUD w. $\mathcal{RFT}$ & 65.80 & 71.90 & 52.35 & 61.05 & 57.95 & 61.81 \\
RCIL-QUD w. $\mathcal{RFT}$ + CEM & 66.60 & 73.05 & 54.85 & 40.85 & 74.90 & 62.05\\
RCIL-QUD w. $\mathcal{RFT}$ + $\mathcal{RTC}$ & 61.10 & 55.95 & 50.20 & 46.95 & 20.05 & 46.86 \\
RCIL-QUD w. $\mathcal{RFT}$ + $\mathcal{RTC}$ + CEM & 59.15 & 55.60 & 45.20 & 31.15 & 33.75 & 44.97 \\
\toprule
\bottomrule
\multirow{2}{*}{Methods} & \multicolumn{6}{c}{CIFAR-100 (SA)} \\ \cmidrule{2-7} 
 & \multicolumn{1}{c|}{$\mathcal{T}_1$ (\%)} & \multicolumn{1}{c|}{$\mathcal{T}_2$ (\%)} & \multicolumn{1}{c|}{$\mathcal{T}_3$ (\%)} & \multicolumn{1}{c|}{$\mathcal{T}_4$ (\%)} & \multicolumn{1}{c|}{$\mathcal{T}_5$ (\%)} & \multicolumn{1}{c}{Average (\%)}\\ \midrule
MTAT$_{\mathrm{lower}}$ & 15.15 & 10.95 & 11.60 & 12.20 & 16.70 & 13.32\\
MTAT$_{\mathrm{upper}}$ & 55.18 & 49.40 & 58.65 & 51.70 & 61.50 & 55.41\\
\midrule
RCIL-QUD w. $\mathcal{RKD}$ & 26.55 & 19.65  & 39.05 & 42.80 & 58.40 & 37.29 \\
RCIL-QUD w. $\mathcal{RKD}$ +CEM & 30.10 & 21.20 & 33.95 & 33.85 & 41.50 & 32.12\\
RCIL-QUD w. $\mathcal{RFT}$ & 17.25 & 16.55 & 27.80 & 28.40 & 68.40 & 31.68 \\
RCIL-QUD w. $\mathcal{RFT}$ + CEM & 27.45 & 21.50 & 31.55 & 27.15 & 43.30 & 30.19\\
\bottomrule
\end{tabular}}
\end{threeparttable}
\end{center}
\end{table}

\subsection{More Results of CIL-QUD with Different Backbones}
We choose another classic model backbone, i.e., ResNet-32, in CIL literature~\citep{zhang2019class,rebuffi2017icarl}. New experiments are conducted on CIFAR-100 and learn 5 and 10 tasks in an incremental manner. Other settings are the same as the one in Table~\ref{table:standard}. We obtain:
\begin{itemize}
    \item Incrementally learning 5 task: DMC v.s. Our CIL-QUD w. $\mathcal{KD}$ = $46.32\%$ v.s. $48.07\%$
    \item Incrementally learning 10 tasks: DMC v.s. Our CIL-QUD w. $\mathcal{KD}$ = $35.99\%$ v.s. $40.16\%$
\end{itemize}
The consistent performance improvements on the new model validate the generalization capability of our proposals.

\subsection{More Results of CIL-QUD with More Tasks}

We conduct additional experiments on CIFAR-100 by incrementally learning $10$ classes at a time (i.e., a higher task number of $10$). Results are collected in Table~\ref{table:10tasks}, where settings are the same as the ones in Table~\ref{table:standard}. We see our proposed methods maintain superior performance (at least $7.91\%$ accuracy gains), compared to previous approaches. Furthermore, we implement DMC~\citep{zhang2019class} and CIL-QUD w. $\mathcal{KD}$ on CIFAR-100 by continually learning $5$ classes at a time (i.e., a higher task number of $20$), where DMC v.s. CIL-QUD w. $\mathcal{KD}$ is $23.74\%$ v.s. $33.13\%$. We see the conclusions are consistent. 

\begin{table}[!ht]
\begin{center}
\caption{The performance of CIL-QUD with $10$ sequentially arrived tasks on CIFAR-100.}
\vspace{-3mm}
\label{table:10tasks}
\begin{threeparttable}
\resizebox{1\textwidth}{!}{
\begin{tabular}{l|c|c|c|c|c|c|c|c|c|c|c}
\toprule
\multirow{2}{*}{Methods} & \multicolumn{11}{c}{CIFAR-100 (SA)} \\ \cmidrule{2-12} 
 & $\mathcal{T}_1$ (\%) & $\mathcal{T}_2$ (\%) & $\mathcal{T}_3$ (\%) & $\mathcal{T}_4$ (\%) & $\mathcal{T}_5$ (\%) & $\mathcal{T}_6$ (\%) & $\mathcal{T}_7$ (\%) & $\mathcal{T}_8$ (\%) & $\mathcal{T}_9$ (\%) & $\mathcal{T}_{10}$ (\%) & Average (\%)\\ \midrule
iCaRL~\citep{rebuffi2017icarl} & $5.90$ & $7.50$ & $4.50$ & $2.80$ & $9.00$ & $8.00$ & $28.20$ & $38.50$ & $59.60$ & $80.20$ & $24.42$ \\
IL2M~\citep{Belouadah_2019_ICCV} & $19.90$ & $24.10$ & $19.80$ & $12.90$ & $21.30$ & $21.70$ & $29.90$ & $34.80$ & $40.30$ & $89.80$ & $31.45$\\ \midrule
Baseline + Auxiliary Classifier & $21.20$ & $32.10$ & $23.00$ & $22.70$ & $21.70$ & $31.70$ & $39.60$ & $33.80$ & $40.30$ & $54.30$ & $32.02$\\
CIL-QUD w. $\mathcal{KD}$ & $29.04$ & $33.94$ & $32.54$ & $27.94$ & $32.74$& $29.64$ & $47.94$ & $45.34$ & $47.24$ & $67.24$ & $39.36$ \\
\bottomrule
\end{tabular}}
\end{threeparttable}
\end{center}
\end{table}




\end{document}

%% file: math_commands.tex
\usepackage{amsmath,amsfonts,bm}

\def\eqref#1{equation~\ref{#1}}

\def\1{\bm{1}}

\DeclareMathAlphabet{\mathsfit}{\encodingdefault}{\sfdefault}{m}{sl}
\SetMathAlphabet{\mathsfit}{bold}{\encodingdefault}{\sfdefault}{bx}{n}

\DeclareMathOperator*{\argmax}{arg\,max}

%% file: AdvCL.bbl
\begin{thebibliography}{63}
\providecommand{\natexlab}[1]{#1}
\providecommand{\url}[1]{\texttt{#1}}
\expandafter\ifx\csname urlstyle\endcsname\relax
  \providecommand{\doi}[1]{doi: #1}\else
  \providecommand{\doi}{doi: \begingroup \urlstyle{rm}\Url}\fi

\bibitem[Alayrac et~al.(2019)Alayrac, Uesato, Huang, Fawzi, Stanforth, and
  Kohli]{alayrac2019labels}
Jean-Baptiste Alayrac, Jonathan Uesato, Po-Sen Huang, Alhussein Fawzi, Robert
  Stanforth, and Pushmeet Kohli.
\newblock Are labels required for improving adversarial robustness?
\newblock In \emph{Advances in Neural Information Processing Systems}, pp.\
  12214--12223, 2019.

\bibitem[Aljundi et~al.(2017)Aljundi, Chakravarty, and
  Tuytelaars]{aljundi2017expert}
Rahaf Aljundi, Punarjay Chakravarty, and Tinne Tuytelaars.
\newblock Expert gate: Lifelong learning with a network of experts.
\newblock In \emph{Proceedings of the IEEE Conference on Computer Vision and
  Pattern Recognition}, pp.\  3366--3375, 2017.

\bibitem[Bateni et~al.(2022)Bateni, Barber, Goyal, Masrani, van~de Meent,
  Sigal, and Wood]{bateni2022beyond}
Peyman Bateni, Jarred Barber, Raghav Goyal, Vaden Masrani, Jan-Willem van~de
  Meent, Leonid Sigal, and Frank Wood.
\newblock Beyond simple meta-learning: Multi-purpose models for multi-domain,
  active and continual few-shot learning.
\newblock \emph{arXiv preprint arXiv:2201.05151}, 2022.

\bibitem[Belouadah \& Popescu(2018)Belouadah and Popescu]{belouadah2018deesil}
Eden Belouadah and Adrian Popescu.
\newblock Deesil: Deep-shallow incremental learning.
\newblock In \emph{Proceedings of the European Conference on Computer Vision
  (ECCV)}, pp.\  0--0, 2018.

\bibitem[Belouadah \& Popescu(2019)Belouadah and Popescu]{Belouadah_2019_ICCV}
Eden Belouadah and Adrian Popescu.
\newblock Il2m: Class incremental learning with dual memory.
\newblock In \emph{The IEEE International Conference on Computer Vision
  (ICCV)}, October 2019.

\bibitem[Belouadah \& Popescu(2020)Belouadah and Popescu]{belouadah2020scail}
Eden Belouadah and Adrian Popescu.
\newblock Scail: Classifier weights scaling for class incremental learning,
  2020.

\bibitem[Boopathy et~al.(2020)Boopathy, Liu, Zhang, Liu, Chen, Chang, and
  Daniel]{boopathy2020proper}
Akhilan Boopathy, Sijia Liu, Gaoyuan Zhang, Cynthia Liu, Pin-Yu Chen, Shiyu
  Chang, and Luca Daniel.
\newblock Proper network interpretability helps adversarial robustness in
  classification.
\newblock In \emph{International Conference on Machine Learning}, pp.\
  1014--1023. PMLR, 2020.

\bibitem[Buda et~al.(2018)Buda, Maki, and Mazurowski]{buda2018systematic}
Mateusz Buda, Atsuto Maki, and Maciej~A Mazurowski.
\newblock A systematic study of the class imbalance problem in convolutional
  neural networks.
\newblock \emph{Neural Networks}, 106:\penalty0 249--259, 2018.

\bibitem[Castro et~al.(2018)Castro, Mar{\'\i}n-Jim{\'e}nez, Guil, Schmid, and
  Alahari]{castro2018end}
Francisco~M Castro, Manuel~J Mar{\'\i}n-Jim{\'e}nez, Nicol{\'a}s Guil, Cordelia
  Schmid, and Karteek Alahari.
\newblock End-to-end incremental learning.
\newblock In \emph{Proceedings of the European Conference on Computer Vision
  (ECCV)}, pp.\  233--248, 2018.

\bibitem[Chan et~al.(2019)Chan, Tay, and Ong]{chan2019thinks}
Alvin Chan, Yi~Tay, and Yew-Soon Ong.
\newblock What it thinks is important is important: Robustness transfers
  through input gradients, 2019.

\bibitem[Chawla et~al.(2002)Chawla, Bowyer, Hall, and
  Kegelmeyer]{chawla2002smote}
Nitesh~V Chawla, Kevin~W Bowyer, Lawrence~O Hall, and W~Philip Kegelmeyer.
\newblock Smote: synthetic minority over-sampling technique.
\newblock \emph{Journal of artificial intelligence research}, 16:\penalty0
  321--357, 2002.

\bibitem[Chen et~al.(2020{\natexlab{a}})Chen, Liu, Chang, Cheng, Amini, and
  Wang]{chen2020adversarial}
Tianlong Chen, Sijia Liu, Shiyu Chang, Yu~Cheng, Lisa Amini, and Zhangyang
  Wang.
\newblock Adversarial robustness: From self-supervised pre-training to
  fine-tuning.
\newblock In \emph{Proceedings of the IEEE/CVF Conference on Computer Vision
  and Pattern Recognition}, pp.\  699--708, 2020{\natexlab{a}}.

\bibitem[Chen et~al.(2020{\natexlab{b}})Chen, Zhang, Liu, Chang, and
  Wang]{chen2020long}
Tianlong Chen, Zhenyu Zhang, Sijia Liu, Shiyu Chang, and Zhangyang Wang.
\newblock Long live the lottery: The existence of winning tickets in lifelong
  learning.
\newblock In \emph{International Conference on Learning Representations},
  2020{\natexlab{b}}.

\bibitem[Chen et~al.(2020{\natexlab{c}})Chen, Kornblith, Norouzi, and
  Hinton]{chen2020simple}
Ting Chen, Simon Kornblith, Mohammad Norouzi, and Geoffrey Hinton.
\newblock A simple framework for contrastive learning of visual
  representations.
\newblock In \emph{International conference on machine learning}, pp.\
  1597--1607. PMLR, 2020{\natexlab{c}}.

\bibitem[Chen et~al.(2020{\natexlab{d}})Chen, Kornblith, Swersky, Norouzi, and
  Hinton]{chen2020big}
Ting Chen, Simon Kornblith, Kevin Swersky, Mohammad Norouzi, and Geoffrey~E
  Hinton.
\newblock Big self-supervised models are strong semi-supervised learners.
\newblock \emph{Advances in Neural Information Processing Systems}, 33,
  2020{\natexlab{d}}.

\bibitem[Chen et~al.(2020{\natexlab{e}})Chen, Fan, Girshick, and
  He]{chen2020improved}
Xinlei Chen, Haoqi Fan, Ross Girshick, and Kaiming He.
\newblock Improved baselines with momentum contrastive learning.
\newblock \emph{arXiv preprint arXiv:2003.04297}, 2020{\natexlab{e}}.

\bibitem[Chu et~al.(2016)Chu, Madhavan, Beijbom, Hoffman, and
  Darrell]{chu2016best}
Brian Chu, Vashisht Madhavan, Oscar Beijbom, Judy Hoffman, and Trevor Darrell.
\newblock Best practices for fine-tuning visual classifiers to new domains.
\newblock In \emph{European conference on computer vision}, pp.\  435--442.
  Springer, 2016.

\bibitem[Coleman et~al.(2020)Coleman, Yeh, Mussmann, Mirzasoleiman, Bailis,
  Liang, Leskovec, and Zaharia]{coleman2019selection}
Cody Coleman, Christopher Yeh, Stephen Mussmann, Baharan Mirzasoleiman, Peter
  Bailis, Percy Liang, Jure Leskovec, and Matei Zaharia.
\newblock Selection via proxy: Efficient data selection for deep learning.
\newblock In \emph{International Conference on Learning Representations}, 2020.
\newblock URL \url{https://openreview.net/forum?id=HJg2b0VYDr}.

\bibitem[Deng et~al.(2009)Deng, Dong, Socher, Li, Li, and
  Fei-Fei]{deng2009imagenet}
Jia Deng, Wei Dong, Richard Socher, Li-Jia Li, Kai Li, and Li~Fei-Fei.
\newblock Imagenet: A large-scale hierarchical image database.
\newblock In \emph{2009 IEEE conference on computer vision and pattern
  recognition}, pp.\  248--255. Ieee, 2009.

\bibitem[Ding et~al.(2020)Ding, Sharma, Lui, and Huang]{ding2020mma}
Gavin~Weiguang Ding, Yash Sharma, Kry Yik~Chau Lui, and Ruitong Huang.
\newblock {MMA} training: Direct input space margin maximization through
  adversarial training.
\newblock In \emph{International Conference on Learning Representations}, 2020.
\newblock URL \url{https://openreview.net/forum?id=HkeryxBtPB}.

\bibitem[Goldblum et~al.(2020)Goldblum, Fowl, Feizi, and
  Goldstein]{goldblum2019adversarially}
Micah Goldblum, Liam Fowl, Soheil Feizi, and Tom Goldstein.
\newblock Adversarially robust distillation.
\newblock In \emph{Proceedings of the AAAI Conference on Artificial
  Intelligence}, volume~34, pp.\  3996--4003, 2020.

\bibitem[Goodfellow et~al.(2013)Goodfellow, Mirza, Xiao, Courville, and
  Bengio]{goodfellow2013empirical}
Ian~J Goodfellow, Mehdi Mirza, Da~Xiao, Aaron Courville, and Yoshua Bengio.
\newblock An empirical investigation of catastrophic forgetting in
  gradient-based neural networks.
\newblock \emph{arXiv preprint arXiv:1312.6211}, 2013.

\bibitem[Goodfellow et~al.(2014)Goodfellow, Shlens, and
  Szegedy]{goodfellow2014explaining}
Ian~J Goodfellow, Jonathon Shlens, and Christian Szegedy.
\newblock Explaining and harnessing adversarial examples.
\newblock \emph{arXiv preprint arXiv:1412.6572}, 2014.

\bibitem[Haixiang et~al.(2017)Haixiang, Yijing, Shang, Mingyun, Yuanyue, and
  Bing]{haixiang2017learning}
Guo Haixiang, Li~Yijing, Jennifer Shang, Gu~Mingyun, Huang Yuanyue, and Gong
  Bing.
\newblock Learning from class-imbalanced data: Review of methods and
  applications.
\newblock \emph{Expert Systems with Applications}, 73:\penalty0 220--239, 2017.

\bibitem[He et~al.(2018)He, Wang, Shan, and Chen]{he2018exemplar}
Chen He, Ruiping Wang, Shiguang Shan, and Xilin Chen.
\newblock Exemplar-supported generative reproduction for class incremental
  learning.
\newblock In \emph{British Machine Vision Conference}, 2018.

\bibitem[He \& Garcia(2009)He and Garcia]{he2009learning}
Haibo He and Edwardo~A Garcia.
\newblock Learning from imbalanced data.
\newblock \emph{IEEE Transactions on knowledge and data engineering},
  21\penalty0 (9):\penalty0 1263--1284, 2009.

\bibitem[He \& Zhu(2021)He and Zhu]{he2021unsupervised}
Jiangpeng He and Fengqing Zhu.
\newblock Unsupervised continual learning via pseudo labels.
\newblock \emph{arXiv preprint arXiv:2104.07164}, 2021.

\bibitem[He et~al.(2016)He, Zhang, Ren, and Sun]{he2016deep}
Kaiming He, Xiangyu Zhang, Shaoqing Ren, and Jian Sun.
\newblock Deep residual learning for image recognition.
\newblock In \emph{Proceedings of the IEEE conference on computer vision and
  pattern recognition}, pp.\  770--778, 2016.

\bibitem[Hendrycks et~al.(2019)Hendrycks, Lee, and
  Mazeika]{pmlr-v97-hendrycks19a}
Dan Hendrycks, Kimin Lee, and Mantas Mazeika.
\newblock Using pre-training can improve model robustness and uncertainty.
\newblock In \emph{Proceedings of the 36th International Conference on Machine
  Learning}, 09--15 Jun 2019.

\bibitem[Hinton et~al.(2015)Hinton, Vinyals, and Dean]{hinton2015distilling}
Geoffrey Hinton, Oriol Vinyals, and Jeff Dean.
\newblock Distilling the knowledge in a neural network.
\newblock \emph{arXiv preprint arXiv:1503.02531}, 2015.

\bibitem[Javed \& Shafait(2018)Javed and Shafait]{javed2018revisiting}
Khurram Javed and Faisal Shafait.
\newblock Revisiting distillation and incremental classifier learning.
\newblock In \emph{Asian Conference on Computer Vision}, pp.\  3--17. Springer,
  2018.

\bibitem[Kemker \& Kanan(2018)Kemker and Kanan]{kemker2017fearnet}
Ronald Kemker and Christopher Kanan.
\newblock Fearnet: Brain-inspired model for incremental learning.
\newblock In \emph{International Conference on Learning Representations}, 2018.
\newblock URL \url{https://openreview.net/forum?id=SJ1Xmf-Rb}.

\bibitem[Krizhevsky et~al.(2009)]{krizhevsky2009learning}
Alex Krizhevsky et~al.
\newblock Learning multiple layers of features from tiny images.
\newblock 2009.

\bibitem[Kurakin et~al.(2016)Kurakin, Goodfellow, and
  Bengio]{kurakin2016adversarial}
Alexey Kurakin, Ian Goodfellow, and Samy Bengio.
\newblock Adversarial machine learning at scale.
\newblock \emph{arXiv preprint arXiv:1611.01236}, 2016.

\bibitem[Lee et~al.(2019)Lee, Lee, Shin, and Lee]{lee2019overcoming}
Kibok Lee, Kimin Lee, Jinwoo Shin, and Honglak Lee.
\newblock Overcoming catastrophic forgetting with unlabeled data in the wild.
\newblock In \emph{Proceedings of the IEEE/CVF International Conference on
  Computer Vision}, pp.\  312--321, 2019.

\bibitem[Lewis \& Gale(1994)Lewis and Gale]{lewis1994sequential}
David~D Lewis and William~A Gale.
\newblock A sequential algorithm for training text classifiers.
\newblock In \emph{SIGIR’94}, pp.\  3--12. Springer, 1994.

\bibitem[Li et~al.(2020)Li, Dong, and Hu]{li2020incremental}
Huaiyu Li, Weiming Dong, and Bao-Gang Hu.
\newblock Incremental concept learning via online generative memory recall.
\newblock \emph{IEEE Transactions on Neural Networks and Learning Systems},
  2020.

\bibitem[Li \& Hoiem(2017)Li and Hoiem]{li2017learning}
Zhizhong Li and Derek Hoiem.
\newblock Learning without forgetting.
\newblock \emph{IEEE transactions on pattern analysis and machine
  intelligence}, 40\penalty0 (12):\penalty0 2935--2947, 2017.

\bibitem[Madry et~al.(2018)Madry, Makelov, Schmidt, Tsipras, and
  Vladu]{madry2017towards}
Aleksander Madry, Aleksandar Makelov, Ludwig Schmidt, Dimitris Tsipras, and
  Adrian Vladu.
\newblock Towards deep learning models resistant to adversarial attacks.
\newblock In \emph{International Conference on Learning Representations}, 2018.
\newblock URL \url{https://openreview.net/forum?id=rJzIBfZAb}.

\bibitem[Mallya et~al.(2018)Mallya, Davis, and Lazebnik]{mallya2018piggyback}
Arun Mallya, Dillon Davis, and Svetlana Lazebnik.
\newblock Piggyback: Adapting a single network to multiple tasks by learning to
  mask weights.
\newblock In \emph{Proceedings of the European Conference on Computer Vision
  (ECCV)}, pp.\  67--82, 2018.

\bibitem[McCloskey \& Cohen(1989)McCloskey and
  Cohen]{mccloskey1989catastrophic}
Michael McCloskey and Neal~J Cohen.
\newblock Catastrophic interference in connectionist networks: The sequential
  learning problem.
\newblock In \emph{Psychology of learning and motivation}, volume~24, pp.\
  109--165. Elsevier, 1989.

\bibitem[Parisi et~al.(2019)Parisi, Kemker, Part, Kanan, and
  Wermter]{parisi2019continual}
German~I Parisi, Ronald Kemker, Jose~L Part, Christopher Kanan, and Stefan
  Wermter.
\newblock Continual lifelong learning with neural networks: A review.
\newblock \emph{Neural Networks}, 2019.

\bibitem[Qu et~al.(2021)Qu, Rahmani, Xu, Williams, and Liu]{qu2021recent}
Haoxuan Qu, Hossein Rahmani, Li~Xu, Bryan Williams, and Jun Liu.
\newblock Recent advances of continual learning in computer vision: An
  overview.
\newblock \emph{arXiv preprint arXiv:2109.11369}, 2021.

\bibitem[Rebuffi et~al.(2017)Rebuffi, Kolesnikov, Sperl, and
  Lampert]{rebuffi2017icarl}
Sylvestre-Alvise Rebuffi, Alexander Kolesnikov, Georg Sperl, and Christoph~H
  Lampert.
\newblock icarl: Incremental classifier and representation learning.
\newblock In \emph{Proceedings of the IEEE conference on Computer Vision and
  Pattern Recognition}, pp.\  2001--2010, 2017.

\bibitem[Rebuffi et~al.(2018)Rebuffi, Bilen, and Vedaldi]{rebuffi2018efficient}
Sylvestre-Alvise Rebuffi, Hakan Bilen, and Andrea Vedaldi.
\newblock Efficient parametrization of multi-domain deep neural networks.
\newblock In \emph{Proceedings of the IEEE Conference on Computer Vision and
  Pattern Recognition}, pp.\  8119--8127, 2018.

\bibitem[Rony et~al.(2019)Rony, Hafemann, Oliveira, Ayed, Sabourin, and
  Granger]{rony2019decoupling}
J{\'e}r{\^o}me Rony, Luiz~G Hafemann, Luiz~S Oliveira, Ismail~Ben Ayed, Robert
  Sabourin, and Eric Granger.
\newblock Decoupling direction and norm for efficient gradient-based l2
  adversarial attacks and defenses.
\newblock In \emph{IEEE Conference on Computer Vision and Pattern Recognition},
  pp.\  4322--4330, 2019.

\bibitem[Rosenfeld \& Tsotsos(2018)Rosenfeld and
  Tsotsos]{rosenfeld2018incremental}
Amir Rosenfeld and John~K Tsotsos.
\newblock Incremental learning through deep adaptation.
\newblock \emph{IEEE transactions on pattern analysis and machine
  intelligence}, 2018.

\bibitem[Rusu et~al.(2016)Rusu, Rabinowitz, Desjardins, Soyer, Kirkpatrick,
  Kavukcuoglu, Pascanu, and Hadsell]{rusu2016progressive}
Andrei~A Rusu, Neil~C Rabinowitz, Guillaume Desjardins, Hubert Soyer, James
  Kirkpatrick, Koray Kavukcuoglu, Razvan Pascanu, and Raia Hadsell.
\newblock Progressive neural networks.
\newblock \emph{arXiv preprint arXiv:1606.04671}, 2016.

\bibitem[Schmidt et~al.(2018)Schmidt, Santurkar, Tsipras, Talwar, and
  Madry]{schmidt2018adversarially}
Ludwig Schmidt, Shibani Santurkar, Dimitris Tsipras, Kunal Talwar, and
  Aleksander Madry.
\newblock Adversarially robust generalization requires more data.
\newblock In \emph{Advances in Neural Information Processing Systems}, 2018.

\bibitem[Settles(2009)]{settles2009active}
Burr Settles.
\newblock Active learning literature survey.
\newblock 2009.

\bibitem[Shafahi et~al.(2020)Shafahi, Saadatpanah, Zhu, Ghiasi, Studer, Jacobs,
  and Goldstein]{shafahi2019adversarially}
Ali Shafahi, Parsa Saadatpanah, Chen Zhu, Amin Ghiasi, Christoph Studer, David
  Jacobs, and Tom Goldstein.
\newblock Adversarially robust transfer learning.
\newblock In \emph{International Conference on Learning Representations}, 2020.
\newblock URL \url{https://openreview.net/forum?id=ryebG04YvB}.

\bibitem[Shin et~al.(2017)Shin, Lee, Kim, and Kim]{shin2017continual}
Hanul Shin, Jung~Kwon Lee, Jaehong Kim, and Jiwon Kim.
\newblock Continual learning with deep generative replay.
\newblock In \emph{Advances in Neural Information Processing Systems}, pp.\
  2990--2999, 2017.

\bibitem[Sinha et~al.(2018)Sinha, Namkoong, and Duchi]{sinha2017certifiable}
Aman Sinha, Hongseok Namkoong, and John Duchi.
\newblock Certifiable distributional robustness with principled adversarial
  training.
\newblock In \emph{International Conference on Learning Representations}, 2018.
\newblock URL \url{https://openreview.net/forum?id=Hk6kPgZA-}.

\bibitem[Tahir et~al.(2009)Tahir, Kittler, Mikolajczyk, and
  Yan]{tahir2009multiple}
Muhammad~Atif Tahir, Josef Kittler, Krystian Mikolajczyk, and Fei Yan.
\newblock A multiple expert approach to the class imbalance problem using
  inverse random under sampling.
\newblock In \emph{International workshop on multiple classifier systems}, pp.\
   82--91. Springer, 2009.

\bibitem[Torralba et~al.(2008)Torralba, Fergus, and Freeman]{torralba200880}
Antonio Torralba, Rob Fergus, and William~T Freeman.
\newblock 80 million tiny images: A large data set for nonparametric object and
  scene recognition.
\newblock \emph{IEEE transactions on pattern analysis and machine
  intelligence}, 30\penalty0 (11):\penalty0 1958--1970, 2008.

\bibitem[Wang et~al.(2021)Wang, Yang, Li, Hong, Li, and Zhu]{wang2021ordisco}
Liyuan Wang, Kuo Yang, Chongxuan Li, Lanqing Hong, Zhenguo Li, and Jun Zhu.
\newblock Ordisco: Effective and efficient usage of incremental unlabeled data
  for semi-supervised continual learning.
\newblock In \emph{Proceedings of the IEEE/CVF Conference on Computer Vision
  and Pattern Recognition}, pp.\  5383--5392, 2021.

\bibitem[Wang et~al.(2019)Wang, Gan, Yang, Wu, and Yan]{wang2019dynamic}
Yiru Wang, Weihao Gan, Jie Yang, Wei Wu, and Junjie Yan.
\newblock Dynamic curriculum learning for imbalanced data classification.
\newblock In \emph{Proceedings of the IEEE International Conference on Computer
  Vision}, pp.\  5017--5026, 2019.

\bibitem[Wang et~al.(2017)Wang, Ramanan, and Hebert]{wang2017growing}
Yu-Xiong Wang, Deva Ramanan, and Martial Hebert.
\newblock Growing a brain: Fine-tuning by increasing model capacity.
\newblock In \emph{Proceedings of the IEEE Conference on Computer Vision and
  Pattern Recognition}, pp.\  2471--2480, 2017.

\bibitem[Xie et~al.(2020)Xie, Luong, Hovy, and Le]{xie2020self}
Qizhe Xie, Minh-Thang Luong, Eduard Hovy, and Quoc~V Le.
\newblock Self-training with noisy student improves imagenet classification.
\newblock In \emph{Proceedings of the IEEE/CVF Conference on Computer Vision
  and Pattern Recognition}, pp.\  10687--10698, 2020.

\bibitem[Zhai et~al.(2020)Zhai, Cai, He, Dan, He, Hopcroft, and
  Wang]{zhai2019adversarially}
Runtian Zhai, Tianle Cai, Di~He, Chen Dan, Kun He, John~E. Hopcroft, and Liwei
  Wang.
\newblock Adversarially robust generalization just requires more unlabeled
  data, 2020.
\newblock URL \url{https://openreview.net/forum?id=H1gdAC4KDB}.

\bibitem[Zhang et~al.(2019{\natexlab{a}})Zhang, Yu, Jiao, Xing, El~Ghaoui, and
  Jordan]{zhang2019theoretically}
Hongyang Zhang, Yaodong Yu, Jiantao Jiao, Eric Xing, Laurent El~Ghaoui, and
  Michael Jordan.
\newblock Theoretically principled trade-off between robustness and accuracy.
\newblock In \emph{International conference on machine learning}, pp.\
  7472--7482. PMLR, 2019{\natexlab{a}}.

\bibitem[Zhang et~al.(2019{\natexlab{b}})Zhang, Liu, Wang, and
  Shen]{zhang2019balance}
Junjie Zhang, Lingqiao Liu, Peng Wang, and Chunhua Shen.
\newblock To balance or not to balance: An embarrassingly simple approach for
  learning with long-tailed distributions, 2019{\natexlab{b}}.

\bibitem[Zhang et~al.(2020)Zhang, Zhang, Ghosh, Li, Tasci, Heck, Zhang, and
  Kuo]{zhang2019class}
Junting Zhang, Jie Zhang, Shalini Ghosh, Dawei Li, Serafettin Tasci, Larry
  Heck, Heming Zhang, and C-C~Jay Kuo.
\newblock Class-incremental learning via deep model consolidation.
\newblock In \emph{Proceedings of the IEEE/CVF Winter Conference on
  Applications of Computer Vision}, pp.\  1131--1140, 2020.

\end{thebibliography}
